\documentclass{article}

\usepackage[final]{neurips_2024}

\usepackage{amsmath,amsfonts}
\usepackage[utf8]{inputenc} 
\usepackage[T1]{fontenc}    
\usepackage{hyperref}       
\usepackage{url}            
\usepackage{booktabs}       
\usepackage{amsfonts}       
\usepackage{nicefrac}       
\usepackage{microtype}      
\usepackage{xcolor}         
\usepackage{graphicx}
\usepackage{amsmath,amsfonts}
\usepackage{algorithmic}
\usepackage{algorithm}
\usepackage{array}
\usepackage{textcomp}
\usepackage{stfloats}
\usepackage{url}
\usepackage{verbatim}
\usepackage{graphicx}
\usepackage{booktabs}
\usepackage{multirow}
\usepackage{authblk}
\usepackage{threeparttable}
\usepackage{graphicx}
\usepackage{orcidlink}
\usepackage{threeparttable}
\usepackage{subcaption}
\usepackage[normalem]{ulem}
\useunder{\uline}{\ul}{}
\usepackage{wrapfig}
\usepackage{wrapfig,lipsum}
\usepackage{hyperref}

\title{Ada-MSHyper: Adaptive Multi-Scale Hypergraph Transformer for Time Series Forecasting}

\author{%
  Zongjiang Shang, Ling Chen\thanks{Corresponding author: Ling Chen.}, Binqing Wu, Dongliang Cui \\
  State Key Laboratory of Blockchain and Data Security\\
  College of Computer Science and Technology\\
  Zhejiang University\\
  \texttt{\{zongjiangshang, lingchen, binqingwu, runnercdl\}@cs.zju.edu.cn} \\
}

\begin{document}

\maketitle

\begin{abstract}
 Although transformer-based methods have achieved great success in multi-scale temporal pattern interaction modeling, two key challenges limit their further development: (1) Individual time points contain less semantic information, and leveraging attention to model pair-wise interactions may cause the information utilization bottleneck. (2) Multiple inherent temporal variations (e.g., rising, falling, and fluctuating) entangled in temporal patterns. To this end, we propose $\bold{\underline{Ada}}$ptive $\bold{\underline{M}}$ulti-$\bold{\underline{S}}$cale $\bold{\underline{Hyper}}$graph Transformer (Ada-MSHyper) for time series forecasting. Specifically, an adaptive hypergraph learning module is designed to provide foundations for modeling group-wise interactions, then a multi-scale interaction module is introduced to promote more comprehensive pattern interactions at different scales. In addition, a node and hyperedge constraint mechanism is introduced to cluster nodes with similar semantic information and differentiate the temporal variations within each scales. Extensive experiments on 11 real-world datasets demonstrate that Ada-MSHyper achieves state-of-the-art performance, reducing prediction errors by an average of 4.56\%, 10.38\%, and 4.97\% in MSE for long-range, short-range, and ultra-long-range time series forecasting, respectively. Code is available at \href{https://github.com/shangzongjiang/Ada-MSHyper}{https://github.com/shangzongjiang/Ada-MSHyper}.
\end{abstract}

\section{Introduction}

Time series forecasting has demonstrated its wide applications across many fields \cite{Autoformer, Informer}, e.g., energy consumption planning, traffic and economics prediction, and disease propagation forecasting. In these real-world applications, the observed time series often demonstrate complex and diverse temporal patterns at different scales\cite{SNAS4MTF,TPRNN,WeatherGNN}. For example, due to periodic human activities, traffic occupation and electricity consumption show clear daily patterns (e.g., afternoon or evening), weekly patterns (e.g., weekday or weekend), and even monthly patterns (e.g., summer or winter). 

Recently, deep models have achieved great success in time series forecasting. To tackle intricate temporal patterns and their interactions at different scales, numerous foundational backbones have emerged, including recurrent neural networks (RNNs) \cite{multiRNN1, TPRNN, multiRNN2}, convolutional neural networks (CNNs) \cite{cnn1,cnn2}, graph neural networks (GNNs) \cite{MSGNet, MAGNN}, and transformers \cite{pyraformer, FEDformer}. Particularly, due to the capabilities of depicting pair-wise interactions and extracting multi-scale representations in sequences, transformers are widely used in time series forecasting. However, some recent studies show that even simple multi-scale MLP \cite{linear1, Dlinear} or na\"ive series decomposition methods \cite{naive2, naive1} can outperform transformer-based methods on various benchmarks. We argue the challenges that limit the effectiveness of transformers in time series forecasting are as follows.

The first one is \textit{semantic information sparsity}. Different from natural language processing (NLP) and computer vision (CV), individual time point in time series contains less semantic information \cite{Inparformer, timesnet}. Compared to pair-wise interactions, group-wise interactions among time points with similar semantic information (e.g., neighboring time points or distant but strongly correlated time points) are more emphasized in time series forecasting. To address the problem of semantic information sparsity, some recent works employ patch-based approaches \cite{tsmixer, PatchTST} and hypergraph structures \cite{MSHyper} to enhance locality and capture group-wise interactions. However, simple partitioning of patches and predefined hypergraph structures may introduce a large amount of noise and be hard to discover implicit interactions.

The second one is \textit{temporal variations entanglement}. Due to the complexity and non-stationary of real-world time series, the temporal patterns of observed time series often contain a large number of inherent variations (e.g., rising, falling, and fluctuating), which may mix and overlap with each other. Especially when there are distinct temporal patterns at different scales, multiple temporal variations are deeply entangled, bringing extreme challenges for time series forecasting. To tackle the problem of temporal variations entanglement, recent studies employ series decomposition \cite{Autoformer, FEDformer} and multi-periodicity analysis \cite{timemixer, timesnet} to differentiate temporal variations at different scales. However, existing methods lack the ability to differentiate temporal variations within each scale, making temporal variations within each scale overlap and become entangled with redundant information.

Motivated by the above, we propose Ada-MSHyper, an $\bold{\underline{Ada}}$ptive $\bold{\underline{M}}$ulti-$\bold{\underline{S}}$cale $\bold{\underline{Hyper}}$graph Transformer for time series forecasting. Specifically, Ada-MSHyper map the input sequence into multi-scale feature representations, then by treating the multi-scale feature representations as nodes, an adaptive multi-scale hypergraph structure is introduced to discover the abundant and implicit group-wise node interactions at different scales. To the best of our knowledge, Ada-MSHyper is the first work that incorporates adaptive hypergraph modeling into time series forecasting. The main contributions are summarized as follows: 

\begin{itemize}
\item{We design an adaptive hypergraph learning (AHL) module to model the abundant and implicit group-wise node interactions at different scales and a multi-scale interaction module to perform hypergraph convolution attention, which empower transformers with the ability to model group-wise pattern interactions at different scales.}

\item{We introduce a node and hyperedge constraint (NHC) mechanism during hypergraph learning phase, which utilizes semantic similarity to cluster nodes with similar semantic information and leverages distance similarity to differentiate the temporal variations within each scales.}

\item {We conduct extensive experiments on 11 real-world datasets. The experimental results demonstrate that Ada-MSHyper achieves state-of-the-art (SOTA) performance, reducing error by an average of 4.56\%, 10.38\%, and 4.97\% in MSE for long-range, short-range, and ultra-long-range time series forecasting, respectively, compared to the best baseline.}
\end{itemize}

\section{Related Work}
\textbf{Deep Models for Time Series Forecasting.} Deep models have shown promising results in time series forecasting. To model temporal patterns at different scales and their interactions, a large number of specially designed backbones have emerged. TAMS-RNNs \cite{multiRNN2} captures periodic temporal dependencies through multi-scale recurrent structures with different update frequencies. TimesNet \cite{timesnet} extends the 1D time series into the 2D space, and models multi-scale temporal pattern interactions through 2D convolution inception blocks. Benefiting from the attention mechanism, transformers have gone beyond contemporaneous RNN- and CNN-based methods and achieved promising results in time series forecasting. FEDformer \cite{FEDformer} combines mixture of expert and frequency attention to capture multi-scale temporal dependencies. Pyraformer \cite{pyraformer} extends the input sequence into multi-scale representations and models the interactions between nodes at different scales through pyramidal attention. Nevertheless, with the rapid emergence of linear forecasters \cite{koopa, Dlinear, linear1}, the effectiveness of transformers in this direction is being questioned. 

Recently, some methods have attempted to fully utilize transformers and paid attention to the inherent properties of time series. Some of these methods are dedicated to addressing the problem of semantic information sparsity in time series forecasting. PatchTST \cite{PatchTST} segments the input sequence into subseries-level patches to enhance locality and capture group-wise interactions. MSHyper \cite{MSHyper} models group-wise interactions through multi-scale hypergraph structures and introduces $k$-hop connections to aggregate information from different range of neighbors. However, constrained by the fixed windows and predefined rules, these methods cannot discover implicit interactions. Others emphasize on addressing the problem of temporal variations entanglement in time series forecasting. 
FilM \cite{film} differentiates temporal variations at different scales by decomposing the input series into different period lengths. iTransformer \cite{Itransformer} combines inverted structures with transformer to learn entangled global temporal variations. However, these methods cannot differentiate temporal variations within each scale, making temporal variations within each scale overlap and become entangled with redundant information.

\textbf{Hypergraph Neural Networks.} 
As a generalized form of GNNs, hypergraph neural networks (HGNNs) have been applied in different fields, e.g., video object segmentation \cite{videohyp}, stock selection \cite{STHAN-SR}, multi-agent trajectory prediction \cite{groupnet}, and time series forecasting \cite{MSHyper}. HyperGCN \cite{hypergcn} is the first work that incorporates convolution operation into hypergraphs, which demonstrates the superiority of HGNNs over ordinary GNNs in capture group-wise interactions. Recent studies \cite{HAHC, MGH} show that HGNNs are promising to model group-wise pattern interactions. LBSN2Vec++ \cite{lbsn2vec} uses hypergraphs for location-based social networks, which leverages heterogeneous hypergraph embeddings to capture mobility and social relationship pattern interactions. GroupNet \cite{groupnet} utilizes multi-scale hypergraph for trajectory prediction, which combines relational reasoning with hypergraph structures to capture group-wise pattern interactions among multiple agents. 

Considering the capability of HGNNs in modeling group-wise interactions, in this work, an adaptive multi-scale hypergraph transformer framework is proposed to model the group-wise pattern interactions at different scales. Specifically, an AHL model is designed to model the abundant and implicit group-wise node interactions. In addition, a NHC mechanism is introduced to cluster nodes with similar semantic information and differentiate temporal variations within each scale, respectively.

\section{Preliminaries}
\textbf{Hypergraph.} A hypergraph is defined as $\mathcal{G}=\{\mathcal{V},\mathcal{E}\}$, where $\mathcal{E}=\{e_1,\ldots,e_m,\ldots,e_M\}$ is the hyperedge set and $\mathcal{V}=\{v_1,\ldots,v_n,\ldots,v_N\}$ is the node set. Each hyperedge represents group-wise interactions by connecting a set of nodes $\{v_1,v_2,\ldots,v_n\}\subseteq\mathcal{V}$. The topology of hypergraph can be represented as an incidence matrix $\mathbf{H}\in\mathbb{R}^{N\times M}$, with entries $\mathbf{H}_{nm}$ defined as follows:

\begin{equation}
\mathbf{H}_{nm}=\left\{\begin{array}{cc}1,&\quad v_n\in e_m\\0,&\quad v_n\notin e_m\end{array}\right.
\end{equation}
The degree of the $n$th node is defined as $d(v_n)=\sum_{m=1}^M\mathbf{H}_{nm}$ and the degree of the $m$th hyperedge is defined as $d(v_m)=\sum_{n=1}^N\mathbf{H}_{nm}$. Further, the node degrees and hyperedge degrees are sorted in diagonal matirces $\textbf{D}_\text{v}\in\mathbb{R}^{N\times N}$ and $\mathbf{D}_\mathrm{e}\in\mathbb{R}^{M\times M}$, respectively.

\textbf{Problem Formulation.} Given the input sequence $\mathbf{X}_{1: T}^{\text{I}}=\left\{\boldsymbol{x}_t \mid \boldsymbol{x}_t \in \mathbb{R}^D, t \in[1, T]\right\}$, where $\boldsymbol{x}_t $ represents the values at time step $t$, $T$ is the input length, and $D$ is the feature dimension. The task of time series forecasting is to predict the future $H$ steps, which can be formulated as follows:
\begin{equation}
\widehat{\mathbf{X}}_{T+1: T+H}^{\text{O}}=\mathcal{F}\left(\mathbf{X}_{1: T}^{\text{I}} ; \theta\right) \in \mathbb{R}^{H \times D},
\end{equation}
where $\widehat{\mathbf X}_{T+1:T+H}^\text{O}$ denotes the forecasting results, $\mathcal{F}$ denotes the mapping function, and $\theta$ denotes the learnable parameters of $\mathcal{F}$. The description of the key notations are given in Appendix \ref{notations}.

\section{Ada-MSHyper}

\begin{figure*}[]
\includegraphics[width=5in]{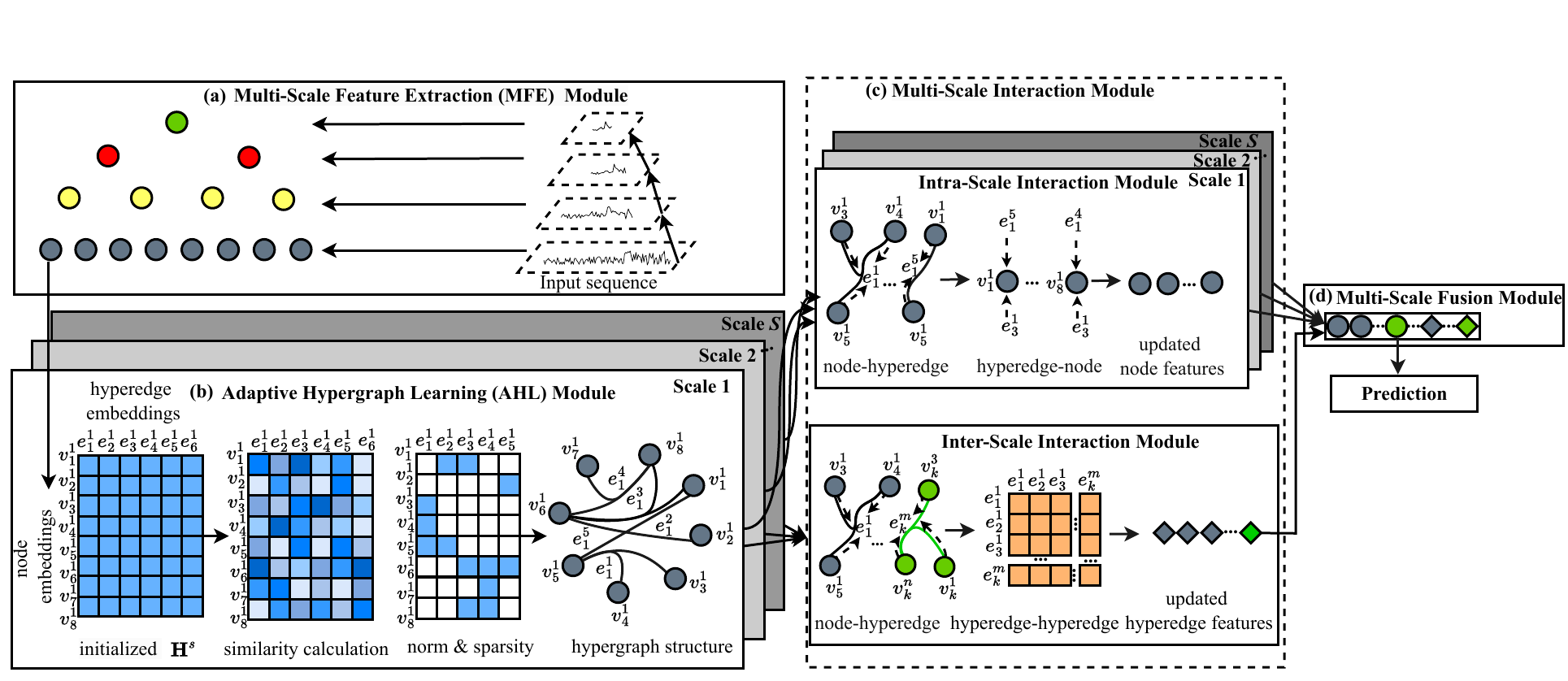}
\centering
\caption{The framework of Ada-MSHyper.}
\label{Figure_1}
\end{figure*}

As previously mentioned, the core of Ada-MSHyper is to promote more comprehensive pattern interactions at different scales. To accomplish this goal, we first map the input sequence into sub-sequences at different scales through the multi-scale feature extraction (MFE) module. Then, by treating multi-scale feature representations as nodes, the AHL module is introduced to model the abundant and implicit group-wise node interactions at different scales. Finally, the multi-scale interaction module is introduced to model group-wise pattern interactions at different scales. Notably, during the hypergraph learning phase, an NHC mechanism is introduced to cluster nodes with similar semantic information and differentiate temporal variations within each scale. The overall framework of Ada-MSHyper is shown in Figure \ref{Figure_1}.

\subsection{Multi-Scale Feature Extraction (MFE) Module}
The MFE module is designed to get the feature representations at different scales. As shown in Figure \ref{Figure_1}(a), suppose $\mathbf{X}^s=\{\boldsymbol{x}_t^s|\boldsymbol{x}_t^s\in\mathbb{R}^D,t\in[1,N^s]\}$ denotes the sub-sequence at scale $s$, where $s=1,...,S$ denotes the scale index and $S$ is the total number of scales. $N^{s}=\left\lfloor\frac{N^{s-1}}{l^{s-1}}\right\rfloor$ is the number of nodes at scale $s$ and $l^{s-1}$ denotes the size of the aggregation window at scale $s-1$. $\mathbf{X}^1=\mathbf{X}^\text{I}_{1:T}$ is the raw input sequence and the aggregation process can be formulated as follows:
\begin{equation}
\mathbf{X}^{s}=Agg(\mathbf{X}^{s-1};\theta^{s-1})\in\mathbb{R}^{N^{s}\times D}, s\geq2,
\end{equation}
where $Agg$ is the aggregation function, e.g., 1D convolution or average pooling, and $\theta^{s-1}$ denotes the learnable parameters of the aggregation function at scale $s-1$.
\subsection{Adaptive Hypergraph Learning (AHL) Module}
The AHL module automatically generates incidence matrices to model implicit group-wise node interactions at different scales. As shown in Figure \ref{Figure_1}(b), we first initialize two kinds of parameters, i.e., node embeddings $\boldsymbol{E}_{\mathrm{node}}^s\in \mathbb{R}^{N^s\times D}$ and hyperedge embeddings $\boldsymbol{E}_{\mathrm{hyper}}^{s}\in \mathbb{R}^{M^s\times D}$ at scale $s$, where $M^s$ is hyperparameters, representing the number of hyperedges at scale $s$. Then, we can obtain the scale-specific incidence matrix $\mathbf{H}^s$ by similarity calculation, which can be formulated as follows:
\begin{equation}
\mathbf{H}^s=SoftMax(ReLU(\boldsymbol{E}_{\mathrm{node}}^s(\boldsymbol{E}_{\mathrm{hyper}}^s)^T)),
\end{equation}
where the $ReLU$ activation function is used to eliminate weak connections and the $SoftMax$ function is applied to normalize the value of $\mathbf{H}^s$. In order to reduce subsequent computational costs and noise interference, the following strategy is designed to sparsify the incidence matrix:
\begin{equation}
\mathbf{H}_{nm}^s=\{\begin{matrix}\mathbf{H}_{nm}^s,&\mathbf{H}_{nm}^s\in TopK(\mathbf{H}_{n*}^s,\eta)\\0,&\mathbf{H}_{nm}^s\notin TopK(\mathbf{H}_{n*}^s,\eta)\end{matrix}
\end{equation}
where $\eta$ is the threshold of $TopK$ function and denotes the max number of neighboring hyperedges connected to a node. The final values of $\mathbf{H}_{nm}^{s}$ can be obtained as follows:
\begin{equation}
\mathbf{H}_{nm}^{s}=\{_{0,\quad\mathbf{H}_{nm}^{s}<\beta}^{1,\quad\mathbf{H}_{nm}^{s}>\beta}
\end{equation}
where $\beta$ denotes the threshold, and the final scale-specific incidence matrices can be represented as $\{\mathbf{H}^1,\cdotp\cdotp\cdotp,\mathbf{H}^s,\cdotp\cdotp\cdotp,\mathbf{H}^S\}$.
Compared to previous methods, our adaptive hypergraph learning is novel from two aspects. Firstly, our methods can capture group-wise interactions at different scales, while most previous methods \cite{Inparformer, Itransformer} can only model pair-wise interactions at a single scale. Secondly, our methods can model abundant and implicit interactions, while many previous methods \cite{PatchTST, MSHyper} depend on fixed windows and predefined rules.

\subsection{Node and Hyperedge Constraint (NHC) Mechanism}  
Although the AHL module can help discover implicit group-wise node interactions at different scales, we argue that the pure data-driven approach faces two limitations, i.e., unable to efficiently cluster nodes with similar semantic information and differentiate temporal variations within each scale. To tackle the above dilemmas, we introduce the NHC mechanism during hypergraph learning phase. 

Given the multi-scale feature representations $\{\mathbf{X}^1,\cdotp\cdotp\cdotp,\mathbf{X}^s,\cdotp\cdotp\cdotp,\mathbf{X}^S\}$ generated from the MFE module, and the scale-specific incidence matrices $\{\mathbf{H}^1,\cdotp\cdotp\cdotp,\mathbf{H}^s,\cdotp\cdotp\cdotp,\mathbf{H}^S\}$ generated from the AHL module, we first get the initialized node feature representations $\boldsymbol{\mathcal{V}}^s=f(\mathbf{X}^s)$ at scale $s$, where $f$ can be implemented by the multi-layer perceptron (MLP). As shown in Figure \ref{Figure_2}(a), the initialized hyperedge feature representations can be obtained by the aggregation operation based on $\mathbf{H}^s$. Specifically, for the $i$th hyperedge $e_i^s$ at scale $s$, its feature representations $\boldsymbol{e}_i^{s}$ can be formulated as follows:
\begin{equation}
\boldsymbol{e}_i^{s}=avg(\sum\nolimits_{v_{j}^s\in\mathcal{N}(e_{i}^s)}\boldsymbol{v}_{j}^{s}){\in\mathbb{R}^{D}},
\end{equation}
where $avg$ is the average operation, $\mathcal{N}(e_{i})$ represents the neighboring nodes connected by $e_i^s$ at scale $s$, and $\boldsymbol{v}_{j}^{s}\in\boldsymbol{\mathcal{V}}^s$ is the $j$th node feature representations at scale $s$. The initialized hyperedge feature representations at different scales can be represented as $\{\boldsymbol{\mathcal{E}}^1,\cdotp\cdotp\cdotp,\boldsymbol{\mathcal{E}}^s,\cdotp\cdotp\cdotp,\boldsymbol{\mathcal{E}}^S\}$. Then, based on semantic similarity and distance similarity, we introduce node constraint to cluster nodes with similar semantic information and leverage hyperedge constraint to differentiate the temporal variations of temporal patterns.

\textbf{Node Constraint.} 
In the data-driven hypergraph, we observe that some nodes connected by the same hyperedge contain distinct semantic information. To cluster nodes with similar semantic information and reduce irrelevant noise interference, we introduce node constraint based on the semantic similarity between nodes and their corresponding hyperedges. As shown in Figure \ref{Figure_2}(b), for the $j$th node at scale $s$, we first obtain its semantic similarity difference $\widetilde{\boldsymbol{v}_{j}^s}$ with its corresponding hyperedges:
\begin{figure*}[]
\includegraphics[width=4in]{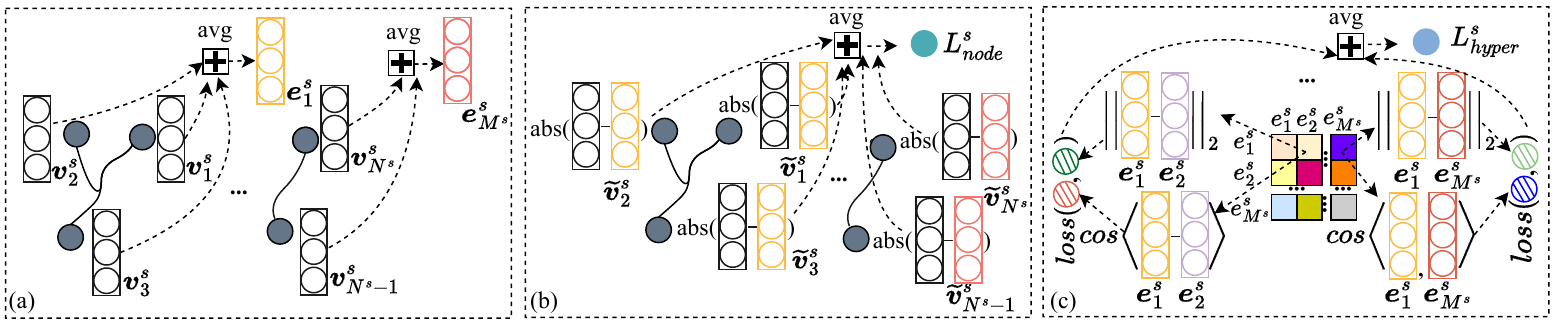}
\centering
\caption{The node and hyperedge constraint mechanism.}
\label{Figure_2}
\end{figure*}

\begin{equation}
\widetilde{\boldsymbol{v}_{j}^s}=\{abs(\boldsymbol{v}_{j}^{s}-\boldsymbol{e}_{{i}}^{s})|v_j^s\in\mathcal{N}(e_i^s)\},
\end{equation}
where $abs$ refers to the operation of calculating the absolute value. The node loss $L_{node}^s$ at scale $s$ based on node constraint can be formulated as follows:
\begin{equation}
L_{node}^s=\frac{1}{N^s}\sum_{i=1}^{N^s}\widetilde{\boldsymbol{v}_{j}^s},
\end{equation}
where ${N^s}$ is the number of nodes at scale $s$. Empowered by the node constraint, our method can enjoy more advantageous group-wise semantic information than pure data-driven hypergraph. Further experimental results and visualization analysis in Section \ref{ablation studies} and Appendix \ref{case appedix} demonstrate the effectiveness of the node constraint in clustering nodes with similar semantic information.

\textbf{Hyperedge Constraint.} Since time series is a collection of data points arranged in chronological order, some recent works \cite{PatchTST, MSHyper} show that connecting multiple nodes sequentially through patches or hyperedges can represent specific temporal variations. Therefore, to deal with the problem of temporal variations entanglement, we introduce hyperedge constraint based on distance similarity. As shown in Figure \ref{Figure_2}(c), we first compute the cosine similarity to reflect the correlation of any two hyperedge representations at scale $s$, which can be formulated as follows: 
\begin{equation}
\alpha_{i,j}=\frac{\boldsymbol{e}_{i}^s(\boldsymbol{e}_{j}^s)^T}{\left\|\boldsymbol{e}_{i}^s\right\|_2\left\|\boldsymbol{e}_{j}^s\right\|_2},
\end{equation}
where $\alpha_{i,j}$ represents the correlation weight. $\boldsymbol{e}_{i}^s$ and $\boldsymbol{e}_{j}^s$ are the $i$th and $j$th hyperedge representation at scale $s$, respectively. Then, we use Euclidean distance $D_{i,j}$ to measure the differentiation magnitude between any two hyperedge representations, which can be formulated as follows:
\begin{equation}
D_{i,j}=\left\|\boldsymbol{e}_{i}^s-\boldsymbol{e}_{j}^s\right\|_2=\sqrt{\sum\nolimits_{d=1}^D((\boldsymbol{e}_{i}^s)^d-(\boldsymbol{e}_{j}^s)^d)^2},
\end{equation}
The hyperedge loss $L_{hyper}^s$ at scale $s$ based on the correlation weight and Euclidean distance can be formulated as follows:
\begin{equation}
L_{hyper}^s=\frac{1}{({M^s})^{2}}\sum\nolimits_{i=1}^{M^s}\sum\nolimits_{j=1}^{M^s}\left(\alpha_{i,j}D_{i,j}+(1-\alpha_{i,j})max(\gamma-D_{i,j},0)\right),
\end{equation}
where $\gamma>0$ denotes the threshold. Notably, when $\alpha_{i,j}=1$, indicating that $e_i^s$ and $e_k^s$ are deemed similar, the hyperedge loss turns to $L_{hyper}=\frac{1}{{(M^s)}^{2}}\sum_{i=1}^{M^s}\sum_{j=1}^{M^s}\alpha_{i,j}D_{i,j}$, where the loss will increase if $D_{i,j}$ becomes large. Conversely, when $\alpha_{i,j}=0$, meaning $e_i$ and $e_k$ are regarded as dissimilar, the hyperedge loss turns to $L_{hyper}=\frac{1}{{(M^s)}^{2}}\sum_{i=1}^{M^s}\sum_{j=1}^{M^s}(1-\alpha_{i,j})max(s-D_{i,j},0)$, where the loss will increase if $D_{i,j}$ falls below the threshold and turns smaller. Other cases lie between the above circumstances. We further provide the visualization results in Section \ref{ablation studies} and appendix \ref{case appedix} to verify that our constraint loss can differentiate temporary variations of temporary patterns within each scale and promote forecasting performance. The final constraint loss $L_{const}$ based on node constraint and hyperedge constraint can be formulated as follows:
\begin{equation}
L_{const}=\lambda\sum\nolimits_{s=1}^SL_{node}^s+(1-\lambda)\sum\nolimits_{s=1}^SL_{hyper}^s,
\end{equation}
wherer $\lambda$ denotes the hyperparameter controlling the balance between node loss and hyperedge loss. 

\subsection{Multi-Scale Interaction Module}
To promote more comprehensive pattern interactions at different scales, a direct way is to mix multi-scale node feature representations at different scales. However, we argue that intra-scale interactions and inter-scale interactions reflect different aspects of pattern interactions, where intra-scale interactions mainly depict detailed interactions between nodes with similar semantic information and inter-scale interactions highlight macroscopic variations interactions\cite{TPRNN,timemixer}. Therefore, instead of directly mixing multi-scale pattern information as a whole, we introduce the multi-scale interaction module to perform inter-scale interactions and intra-scale interactions.

\textbf{Intra-Scale Interaction Module.} 
Due to the semantic information sparsity of time series, traditional pair-wise attention may may cause the information utilization bottleneck \cite{Inparformer}. In contrast, some recent studies \cite{PatchTST, MSHyper} show that group-wise interactions can provide more informative insights in time series forecasting. To capture group-wise interactions among nodes with similar semantic information within each scale, we introduce hypergraph convolution attention within the intra-scale interaction module. Specifically, given $\mathbf{H}^s$, we first use attention mechanism to capture the interaction strength of each node $v_i^s\in\boldsymbol{\mathcal{V}}^s$ and its related hyperedges at scale $s$, which can be formulated as follows:
\begin{equation}
\boldsymbol{\mathcal{H}}^s_{ij}=\frac{\exp(\sigma(f_t[\boldsymbol{v}_i^s,{\boldsymbol{e}}_{j}^s]))}{\sum_{e_k^s\in\mathcal{N}(v_i^s)}exp(\sigma(f_t[\boldsymbol{v}_i^s,{\boldsymbol{e}}_{k}^s])},
\end{equation}
where $[.,.]$ denotes the concatenation operation of the $i$th node and its related hyperedges. $f_t$ is a trainable MLP, and $\mathcal{N}(v_i^s)$ is the neighboring hyperedges connected to $v_i^s$, which can be accessed using ${\mathbf{H}}^s$. Then, considering the symmetric normalized hypergraph Laplacian convolution $\Delta=\mathbf{D}_{v}^{-1/2}\mathbf{H}\mathbf{W}\mathbf{D}_{e}^{-1}\mathbf{H}^{T}\mathbf{D}_{v}^{-1/2}$ used in HGNN \cite{HGNN}, the multi-head hypergraph convolution attention can be formulated as follows:
\begin{equation}
\boldsymbol{\widetilde{\mathcal{V}}}^s=\bigoplus_{\jmath=1}^{\mathcal{J}}(\sigma(\mathbf{D}_{v^s}^{-1/2}\boldsymbol{\mathcal{H}}_{\jmath}^{s}\mathbf{D}_{e^s}^{-1}(\boldsymbol{\mathcal{H}}_\jmath^s)^\mathrm{T}\mathbf{D}_{v^s}^{-1/2}\boldsymbol{\mathcal{V}}^s\mathbf{P}^s_\jmath))\in\mathbb{R}^{N^s\times D},
\end{equation}
 where $\boldsymbol{\widetilde{\mathcal{V}}}^s$ is the updated node feature representations at scale $s$, $\bigoplus$ is the aggregation function used for combing the outputs of multi-head, e.g., concatenation or average pooling. $\sigma$ is the activation function, e.g., LeakyReLU and ELU. $\boldsymbol{\mathcal{H}}^{s}_{\jmath}$ and $\mathbf{P}^s_{\jmath}$ are the enriched incidence matrix and the learnable weight matrix of the $\jmath$th head at scale $s$, respectively. $\mathcal{J}$ is the number of heads. 
 
\textbf{Inter-Scale Interaction Module.}
The inter-scale interaction module is introduced to capture pattern interactions at different scales. To achieve this goal, a direct way is to model group-wise node interactions across all scales. However, detailed group-wise node interactions across all scales can introduce redundant information and increase computation complexity. Therefore, we adopt a hyperedge attention within the inter-scale interaction module to capture macroscopic variations interactions at different scales. Technically, based on the hyperedge representations $\boldsymbol{\mathcal{E}}=\{\boldsymbol{\mathcal{E}}^1,\cdotp\cdotp\cdotp,\boldsymbol{\mathcal{E}}^s,\cdotp\cdotp\cdotp,\boldsymbol{\mathcal{E}}^S\}$, we first adopt linear  projections to get queries, keys,  and values $\mathbf{Q}$,  $\mathbf{K}$,  $\mathbf{V}\in\mathbb{R}^{M\times D}$. Then the hyperedge attention can be formulated as follows:
\begin{equation}
\tilde{\mathbf{V}}=softmax(\frac{\mathbf{QK}^\mathrm{T}}{\sqrt{D_K}})\mathbf{V},
\end{equation}

where $\tilde{\mathbf{V}}$ is the updated hyperedge feature representations. 
\subsection{Prediction Module \& Loss Function}
After obtaining the updated node and hyperedge feature representations, we concatenate them and feed them into a linear layer for prediction. We choose Mean Squared Error (MSE) as our forecasting loss, which can be formulated as follows:
\begin{equation}
 L_{mse}=\frac{1}{H}\left\|\widehat{\mathbf X}_{T+1:T+H}^\text{O}-\mathbf X_{T+1:T+H}^\text{O}\right\|_2^2,
\end{equation}
where $\mathbf X_{T+1:T+H}^\text{O}$ and $\widehat{\mathbf X}_{T+1:T+H}^\text{O}$ are ground truth and forecasting results, respectively. Notably, during training phase, $L_{mse}$ is used to regulate the overall learning process, while $L_{const}$ is only used to constrain hypergraph learning process. 
\subsection{Complexity Analysis}
For the MFE module, the time complexity is $\mathcal{O}(Nl)$, where $N$ is the number of nodes at the finest scale and $N$ is equal to the input length $T$. $l$ is the aggregation window size at the finest scale. For the AHL module, the time complexity is $\mathcal{O}(MN+M^2)$, where $M$ is the number of hypergraphs at the finest scale. For the intra-scale interaction module, since $\mathbf{D}_v$ and $\mathbf{D}_e$ are diagonal matrices, the time complexity is $\mathcal{O}(MN)$. For the inter-scale interaction module, the time complexity is $M^2$. In practical operation, $M$ and $l$ is the hyperparameter and is much smaller than $N$. As a result, the total time complexity of Ada-MSHyper is bounded by $\mathcal{O}(N)$.
\section{Experiment}
\subsection{Experimental Setup}\label{setup}
\begin{wraptable}{r}{0.54 \textwidth}
\centering
  \caption{Dataset statistics.}
  \label{tab:table1}
  \resizebox{ \linewidth}{!}{
    \begin{tabular}{@{}llllll@{}}
    \toprule
    \multicolumn{1}{l}{Dataset} & \# Variates & Prediction Length & Frequency & Forecastability & Information \\ \midrule
    \multicolumn{1}{l|}{ETT (4 subsets)} & \multicolumn{1}{l|}{7} & \multicolumn{1}{l|}{(96, 192, 336,720)} & \multicolumn{1}{l|}{(15 mins, Hourly)} & \multicolumn{1}{l|}{(0.38-0.55)} & Temperature \\ \midrule
    \multicolumn{1}{l|}{Weather} & \multicolumn{1}{l|}{21} & \multicolumn{1}{c|}{(96, 192, 336,720)} & \multicolumn{1}{l|}{10 mins} & \multicolumn{1}{l|}{0.75} & Weather \\ \midrule
    \multicolumn{1}{l|}{Electricity} & \multicolumn{1}{l|}{321} & \multicolumn{1}{c|}{(96, 192, 336,720)} & \multicolumn{1}{l|}{Hourly} & \multicolumn{1}{l|}{0.77} & Electricity \\ \midrule
    \multicolumn{1}{l|}{Traffic} & \multicolumn{1}{l|}{862} & \multicolumn{1}{l|}{(96, 192, 336,720)} & \multicolumn{1}{l|}{Hourly} & \multicolumn{1}{l|}{0.68} & Transportation \\ \midrule
    \multicolumn{1}{l|}{PEMS (4 subsets)} & \multicolumn{1}{l|}{(170-883)} & \multicolumn{1}{l|}{(12, 24, 48)} & \multicolumn{1}{l|}{5 mins} & \multicolumn{1}{l|}{(0.43-0.58)} & Traffic network \\ \bottomrule
    \end{tabular}
    }
\end{wraptable}
\textbf{Datasets.} For long-range time series forecasting, we conduct experiments on 7 commonly used benchmarks, including ETT (ETTh1, ETTh2, ETTm1, and ETTm2), Traffic, Electricity, and Weather datasets following \cite{Autoformer, Itransformer, MSHyper}. For short-range time series forecasting, we adopt 4 benchmarks from PEMS (PEMS03, PEMS04, PEMS07, and PEMS08) following \cite{Itransformer,timemixer}. For ultra-long-range time series forecasting, we adopt ETT datasets following \cite{witran}. Table \ref{tab:table1} gives the dataset statistics. In addition, the forecastability is derived from one minus the entropy of the Fourier decomposition of a time series\cite{timemixer, forecastable}. Higher values mean greater forecastability.

\textbf{Baselines.} We compare Ada-MSHyper with 15 competitive baselines, i.e., iTransformer \cite{Itransformer}, MSHyper \cite{MSHyper}, PatchTST \cite{PatchTST}, TimeMixer \cite{timemixer}, MSGNet \cite{MSGNet}, CrossGNN \cite{CrossGNN}, TimesNet \cite{timesnet}, WITRAN \cite{witran}, SCINet \cite{scinet}, Crossformer \cite{Crossformer}, FiLM \cite{film}, DLinear \cite{Dlinear}, FEDformer \cite{FEDformer}, Pyraformer \cite{pyraformer}, and Autoformer \cite{Autoformer}.

\textbf{Experimental Settings.} Ada-MSHyper is trained/tested on a single NVIDIA Geforce RTX 3090 GPU. MSE and MAE are used as evaluation metrics and lower values mean better performance. Adam is set as the optimizer with the initial learning rate of $\text{10}^{\text{-4}}$. It is notable that the above mentioned baseline results cannot be used directly due to different input and output lengths. For a fair comparison, we set the commonly used input length $T=96$ and output lengths $H\in \{96,192,336,720\}$ for long-range forecasting, $H\in \{12,24,48\}$ for short-range forecasting, and $H\in \{1080, 1440, 1800,2160\}$ for ultra-long-range forecasting. More descriptions about datasets, baselines, and experimental settings are given in Appendix \ref{datasets}, \ref{baselines}, and \ref{experimental settings}, respectively.

\subsection{Main Results}\label{main results}

\textbf{Long-Range Forecasting.} Table \ref{tab:table2} shows the results of long-range time series forecasting under multivariate settings. We can observe that: (1) Ada-MSHyper achieves the SOTA results in all datasets, with an average error reduction of 4.56\% and 3.47\% compared to the best baseline in MSE and MAE, respectively.  (2) FEDformer and Autoformer exhibit relatively poor predictive performance. This may be that vanilla attention and simplistic decomposition techniques are insufficient in capturing multi-scale pattern interactions. (3) By considering multi-scale pattern interactions, TimeMixer achieves competitive results. However, its performance deteriorates on the datasets with low forecastability (e.g., ETTh1 and ETTh2 datasets). In contrast, Ada-MSHyper still maintains superiority on low forecastability datasets by modeling group-wise pattern interactions. Notably, despite modeling group-wise pattern interactions, the performance of MSHyper and PatchTST still lags behind that of Ada-MSHyper, indicating that predefined rules may overlook implicit interactions and introduce noise interference for forecasting. Moreover, for long-range time series forecasting under univariate settings, Ada-MSHyper gives an average error reduction of 7.57\% and 4.65\% compared to the best baseline in MSE and MAE, respectively. The univariate results are given in Appendix \ref{full_res}.

\begin{table}[htbp]
\begin{tiny}
  \centering
  \caption{Results of long-range time series forecasting under multivariate settings. The best results are \textbf{\textcolor{red}{bolded}} and the second best results are {\color[HTML]{00B0F0}{\ul underlined}}. Results are averaged from all prediction lengths. Full results are listed in Appendix \ref{full_res}.}
  \label{tab:table2}
  \resizebox{\linewidth}{!}{
     \begin{tabular}{cccccccccccccc}
    \toprule
    Models & \begin{tabular}[c]{@{}c@{}}Ada-MSHyper\\      (Ours)\end{tabular} & \begin{tabular}[c]{@{}c@{}}iTransformer\\      (2024)\end{tabular} & \begin{tabular}[c]{@{}c@{}}MSHyper\\      (2024)\end{tabular} & \begin{tabular}[c]{@{}c@{}}TimeMixer\\      (2024)\end{tabular} & \begin{tabular}[c]{@{}c@{}}MSGNet\\      (2024)\end{tabular} & \begin{tabular}[c]{@{}c@{}}CrossGNN\\      (2023)\end{tabular} 
    & \begin{tabular}[c]{@{}c@{}}PatchTST\\      (2023)\end{tabular} & \begin{tabular}[c]{@{}c@{}}Crossformer\\      (2023)\end{tabular} & \begin{tabular}[c]{@{}c@{}}TimesNet\\      (2023)\end{tabular} & \begin{tabular}[c]{@{}c@{}}DLinear\\      (2023)\end{tabular} & \begin{tabular}[c]{@{}c@{}}FiLM\\      (2022)\end{tabular} & \begin{tabular}[c]{@{}c@{}}FEDformer\\      (2022)\end{tabular} & \begin{tabular}[c]{@{}c@{}}Autoformer\\      (2021)\end{tabular} \\ \midrule
    \multicolumn{1}{c|}{Metric} & \multicolumn{1}{c|}{MSE MAE} & \multicolumn{1}{c|}{MSE MAE} & \multicolumn{1}{c|}{MSE MAE} & \multicolumn{1}{c|}{MSE MAE} & \multicolumn{1}{c|}{MSE MAE} & \multicolumn{1}{c|}{MSE MAE} & \multicolumn{1}{c|}{MSE MAE} & \multicolumn{1}{c|}{MSE MAE} & \multicolumn{1}{c|}{MSE MAE} & \multicolumn{1}{c|}{MSE MAE} & \multicolumn{1}{c|}{MSE MAE} & \multicolumn{1}{c|}{MSE MAE} & MSE MAE \\ \midrule
    
    \multicolumn{1}{c|}{Weather} & \multicolumn{1}{c|}{{\color[HTML]{FF0000} \textbf{0.233 0.259}}} & \multicolumn{1}{c|}{0.258 0.278} & \multicolumn{1}{c|}{0.250 0.279} & \multicolumn{1}{c|}{{{\color[HTML]{00B0F0}{\ul 0.245}} \color[HTML]{00B0F0}{\ul0.276}}} & \multicolumn{1}{c|}{0.249 0.278} & \multicolumn{1}{c|}{0.247 0.289} & \multicolumn{1}{c|}{0.259 0.281} & \multicolumn{1}{c|}{0.259 0.315} & \multicolumn{1}{c|}{0.259 0.287} & \multicolumn{1}{c|}{0.265 0.317} & \multicolumn{1}{c|}{0.253 0.309} & \multicolumn{1}{c|}{0.309 0.360} & 0.338 0.382 \\ \midrule
    
    \multicolumn{1}{c|}{Electricity} & \multicolumn{1}{c|}{{\color[HTML]{FF0000} \textbf{0.167 0.259}}} & \multicolumn{1}{c|}{{\color[HTML]{00B0F0} {\ul 0.178} \color[HTML]{00B0F0} {\ul0.270}}} & \multicolumn{1}{c|}{0.191 0.283} & \multicolumn{1}{c|}{0.182 0.273} & \multicolumn{1}{c|}{0.194 0.300} & \multicolumn{1}{c|}{0.201 0.300} & \multicolumn{1}{c|}{0.205 0.290} & \multicolumn{1}{c|}{0.244 0.334} & \multicolumn{1}{c|}{0.193 0.295} & \multicolumn{1}{c|}{0.212 0.300} & \multicolumn{1}{c|}{0.223 0.302} & \multicolumn{1}{c|}{0.214 0.327} & 0.227 0.364 \\ \midrule
    
    \multicolumn{1}{c|}{ETTh1} & \multicolumn{1}{c|}{{\color[HTML]{FF0000} \textbf{0.418 0.426}}} & \multicolumn{1}{c|}{0.454 0.448} & \multicolumn{1}{c|}{0.455 0.445} & \multicolumn{1}{c|}{0.460 0.445} & \multicolumn{1}{c|}{0.452 0.452} & \multicolumn{1}{c|}{\color[HTML]{00B0F0} {\ul0.437} \color[HTML]{00B0F0} {\ul0.434}} &\multicolumn{1}{c|}{0.469 0.455} & \multicolumn{1}{c|}{0.529 0.522} & \multicolumn{1}{c|}{0.458 0.450} & \multicolumn{1}{c|}{0.456 0.452} & \multicolumn{1}{c|}{0.516 0.483} & \multicolumn{1}{c|}{ 0.440 0.460} & 0.496 0.487 \\ \midrule
    
    \multicolumn{1}{c|}{ETTh2} & \multicolumn{1}{c|}{{\color[HTML]{FF0000} \textbf{0.371 0.394}}} & \multicolumn{1}{c|}{{\color[HTML]{00B0F0} {\ul 0.383} \color[HTML]{00B0F0} {\ul0.407}}} & \multicolumn{1}{c|}{0.385 0.408} & \multicolumn{1}{c|}{0.393 0.412} & \multicolumn{1}{c|}{0.396 0.417} & \multicolumn{1}{c|}{0.393 0.418} &\multicolumn{1}{c|}{0.387 \color[HTML]{00B0F0} {\ul0.407}} & \multicolumn{1}{c|}{0.942 0.684} & \multicolumn{1}{c|}{0.414 0.427} & \multicolumn{1}{c|}{0.559 0.515} & \multicolumn{1}{c|}{0.402 0.420} & \multicolumn{1}{c|}{0.437 0.449} & 0.450 0.459 \\ \midrule
    
    \multicolumn{1}{c|}{ETTm1} & \multicolumn{1}{c|}{{\color[HTML]{FF0000} \textbf{0.365 0.390}}} & \multicolumn{1}{c|}{0.407 0.410} & \multicolumn{1}{c|}{0.412 0.405} & \multicolumn{1}{c|}{{\color[HTML]{00B0F0} {\ul 0.384} \color[HTML]{00B0F0} {\ul0.397}}} & \multicolumn{1}{c|}{0.398 0.411} & \multicolumn{1}{c|}{0.393 0.404} &\multicolumn{1}{c|}{0.387 0.400} & \multicolumn{1}{c|}{0.513 0.495 } & \multicolumn{1}{c|}{0.400 0.406} & \multicolumn{1}{c|}{0.403 0.407} & \multicolumn{1}{c|}{0.411 0.402} & \multicolumn{1}{c|}{0.448 0.452} & 0.588 0.517 \\ \midrule
    
    \multicolumn{1}{c|}{ETTm2} & \multicolumn{1}{c|}{{\color[HTML]{FF0000} \textbf{0.263 0.322}}} & \multicolumn{1}{c|}{0.288 0.332} & \multicolumn{1}{c|}{0.296 0.336} & \multicolumn{1}{c|}{{\color[HTML]{00B0F0} {\ul 0.278} \color[HTML]{00B0F0} {\ul0.325}}} & \multicolumn{1}{c|}{0.288 0.330} & \multicolumn{1}{c|}{0.282 0.330} &\multicolumn{1}{c|}{0.281 0.326} & \multicolumn{1}{c|}{0.757 0.611} & \multicolumn{1}{c|}{0.291 0.333} & \multicolumn{1}{c|}{0.350 0.401} & \multicolumn{1}{c|}{0.288 0.329} & \multicolumn{1}{c|}{0.305 0.349} & 0.327 0.371 \\ \midrule
    
    \multicolumn{1}{c|}{Traffic} & \multicolumn{1}{c|}{{\color[HTML]{FF0000} \textbf{0.415 0.262}}} & \multicolumn{1}{c|}{{\color[HTML]{00B0F0} {\ul 0.428} \color[HTML]{00B0F0} {\ul0.282}}} & \multicolumn{1}{c|}{0.433 0.283} & \multicolumn{1}{c|}{0.492 0.304} &  \multicolumn{1}{c|}{0.641 0.370} & \multicolumn{1}{c|}{0.583 0.323} &\multicolumn{1}{c|}{0.481 0.304} & \multicolumn{1}{c|}{0.550 0.304} & \multicolumn{1}{c|}{0.620 0.336} & \multicolumn{1}{c|}{0.625 0.383} & \multicolumn{1}{c|}{0.637 0.384} & \multicolumn{1}{c|}{0.610 0.376} & 0.628 0.379 \\ \bottomrule
    \end{tabular}}
  \label{tab:addlabel}
\end{tiny}
\end{table}%

\textbf{Short-Range Forecasting.} Table \ref{tab:table3} summarizes the results of short-range time series forecasting under multivariate settings. 
It is notable that the PEMS datasets record multiple time series of citywide traffic networks and show complex spatial-temporal correlations among multiple variates. We adopt the same settings as iTransformer \cite{MSHyper} and TimeMixer \cite{pyraformer}. Ada-MSHyper still achieves the best performance in PEMS datasets, verifying its effectiveness in handling complex multivariate short-range time series forecasting. Specifically, Ada-MSHyper gives an average error reduction of 10.38\% and 3.82\% compared to the best baseline in terms of MSE and MAE, respectively.

\begin{table}[htbp]
\begin{tiny}
  \centering
  \caption{Results of short-range time series forecasting under multivariate settings. Results are averaged from all prediction lengths. Full results are listed in Appendix \ref{full_res}.}
  \label{tab:table3}
    \resizebox{\linewidth}{!}{
\begin{tabular}{@{}clccccccccccc@{}}
\toprule
\multicolumn{2}{c}{Models} & \begin{tabular}[c]{@{}c@{}}Ada-MSHyper\\ (Ours)\end{tabular} & \begin{tabular}[c]{@{}c@{}}iTransformer\\ (2024)\end{tabular} & \begin{tabular}[c]{@{}c@{}}MSHyper\\ (2024)\end{tabular} & \begin{tabular}[c]{@{}c@{}}TimeMixer\\ (2024)\end{tabular} & \begin{tabular}[c]{@{}c@{}}PatchTST\\      (2023)\end{tabular} & \begin{tabular}[c]{@{}c@{}}TimesNet\\      (2023)\end{tabular} & \begin{tabular}[c]{@{}c@{}}DLinear\\      (2023)\end{tabular} & \begin{tabular}[c]{@{}c@{}}Crossformer\\      (2023)\end{tabular} & \begin{tabular}[c]{@{}c@{}}SCINet\\      (2022)\end{tabular} & \begin{tabular}[c]{@{}c@{}}FEDformer\\      (2022)\end{tabular} & \begin{tabular}[c]{@{}c@{}}Autoformer\\      (2021)\end{tabular} \\ \midrule
\multicolumn{2}{c|}{Metric} & \multicolumn{1}{c|}{MSE MAE} & \multicolumn{1}{c|}{MSE MAE} & \multicolumn{1}{c|}{MSE MAE} & \multicolumn{1}{c|}{MSE MAE} & \multicolumn{1}{c|}{MSE MAE} & \multicolumn{1}{c|}{MSE MAE} & \multicolumn{1}{c|}{MSE MAE} & \multicolumn{1}{c|}{MSE MAE} & \multicolumn{1}{c|}{MSE MAE} & \multicolumn{1}{c|}{MSE MAE} & MSE MAE \\ \midrule
\multicolumn{2}{c|}{PEMS03} & \multicolumn{1}{c|}{{\color[HTML]{FF0000} \textbf{0.085 0.193}}} & \multicolumn{1}{c|}{0.096 0.204} & \multicolumn{1}{c|}{0.123 0.226} & \multicolumn{1}{c|}{0.188 0.361} & \multicolumn{1}{c|}{0.151 0.265} & \multicolumn{1}{c|}{0.119 0.225} & \multicolumn{1}{c|}{0.219 0.328} & \multicolumn{1}{c|}{0.138 0.253} & \multicolumn{1}{c|}{{\color[HTML]{00B0F0}{\ul 0.093} \color[HTML]{00B0F0}{\ul0.203}}} & \multicolumn{1}{c|}{0.167 0.291} & 0.546 0.536 \\ \midrule
\multicolumn{2}{c|}{PEMS04} & \multicolumn{1}{c|}{{\color[HTML]{FF0000} \textbf{0.080 0.189}}} & \multicolumn{1}{c|}{0.098 0.207} & \multicolumn{1}{c|}{0.147 0.250} & \multicolumn{1}{c|}{0.183 0.363} & \multicolumn{1}{c|}{0.162 0.279} & \multicolumn{1}{c|}{0.109 0.220} & \multicolumn{1}{c|}{0.242 0.350} & \multicolumn{1}{c|}{0.145 0.267} & \multicolumn{1}{c|}{{\color[HTML]{00B0F0}{\ul 0.085} \color[HTML]{00B0F0}{\ul0.194}}} & \multicolumn{1}{c|}{0.195 0.308} & 0.510 0.537 \\ \midrule
\multicolumn{2}{c|}{PEMS07} & \multicolumn{1}{c|}{{\color[HTML]{FF0000} \textbf{0.076 0.177}}} & \multicolumn{1}{c|}{{\color[HTML]{00B0F0}{\ul 0.088} \color[HTML]{00B0F0}{\ul0.190}}} & \multicolumn{1}{c|}{0.128 0.234} & \multicolumn{1}{c|}{0.172 0.351} & \multicolumn{1}{c|}{0.166 0.270} & \multicolumn{1}{c|}{0.106 0.208} & \multicolumn{1}{c|}{0.241 0.343} & \multicolumn{1}{c|}{0.181 0.272} & \multicolumn{1}{c|}{0.112 0.211} & \multicolumn{1}{c|}{0.133 0.252} & 0.304 0.409 \\ \midrule
\multicolumn{2}{c|}{PEMS08} & \multicolumn{1}{c|}{{\color[HTML]{FF0000} \textbf{0.110 0.210}}} & \multicolumn{1}{c|}{{\color[HTML]{00B0F0}{\ul 0.127} \color[HTML]{00B0F0}{\ul0.212}}} & \multicolumn{1}{c|}{0.220 0.260} & \multicolumn{1}{c|}{0.189 0.374} & \multicolumn{1}{c|}{0.238 0.289} & \multicolumn{1}{c|}{0.150 0.244} & \multicolumn{1}{c|}{0.281 0.366} & \multicolumn{1}{c|}{0.232 0.276} & \multicolumn{1}{c|}{0.133 0.225} & \multicolumn{1}{c|}{0.234 0.323} & 0.623 0.573 \\ \bottomrule
\end{tabular}
  }
  \end{tiny}
\end{table}%

\textbf{Ultra-Long-Range Forecasting.} Table \ref{tab:table4} summarizes the results of ultra-long-range time series forecasting under multivariate settings. We can see that: (1) Ada-MSHyper achieves SOTA results in almost all benchmarks, with an average error reduction of 4.97\% and 2.21\% compared to the best baseline in MSE and MAE, respectively. (2) Compared with other baselines, PatchTST and MSHyper achieve competitive results. The reason may be that group-wise interactions can help mitigate the issue of semantic information sparsity. (3) Compared to PatchTST and MSHyper, Ada-MSHyper achieves superior performance, the reason may be that the inter-scale interaction module can help capture macroscopic variations interactions, especially for the ultra-long-rang time series.

\begin{table}[htbp]
  \centering
  \caption{Results of ultra-long-range time series forecasting under multivariate settings. Results are averaged from all prediction lengths. Full results are listed in Appendix \ref{full_res}.}
  \label{tab:table4}
  \resizebox{\linewidth}{!}{
    \begin{tabular}{cccccccccccc}
    \toprule
    Models & \begin{tabular}[c]{@{}c@{}}Ada-MSHyper\\ (Ours)\end{tabular} & \begin{tabular}[c]{@{}c@{}}iTransformer\\ (2024)\end{tabular} & \begin{tabular}[c]{@{}c@{}}MSHyper\\ (2024)\end{tabular} & \begin{tabular}[c]{@{}c@{}}TimeMixer\\ (2024)\end{tabular} & \begin{tabular}[c]{@{}c@{}}WITRAN\\ (2023)\end{tabular} & \begin{tabular}[c]{@{}c@{}}PatchTST\\ (2023)\end{tabular} & \begin{tabular}[c]{@{}c@{}}DLinear\\ (2023)\end{tabular} & \begin{tabular}[c]{@{}c@{}}Crossformer\\ (2023)\end{tabular} & \begin{tabular}[c]{@{}c@{}}FEDformer\\ (2022)\end{tabular} & \begin{tabular}[c]{@{}c@{}}Pyraformer\\ (2022)\end{tabular} & \begin{tabular}[c]{@{}c@{}}Autoformer\\ (2021)\end{tabular} \\ \midrule
    \multicolumn{1}{c|}{Metric} & \multicolumn{1}{c|}{MSE MAE} & \multicolumn{1}{c|}{MSE MAE} & \multicolumn{1}{c|}{MSE MAE} & \multicolumn{1}{c|}{MSE MAE} & \multicolumn{1}{c|}{MSE MAE} & \multicolumn{1}{c|}{MSE MAE} & \multicolumn{1}{c|}{MSE MAE} & \multicolumn{1}{c|}{MSE MAE} & \multicolumn{1}{c|}{MSE MAE} & \multicolumn{1}{c|}{MSE MAE} & MSE MAE \\ \midrule
    \multicolumn{1}{c|}{ETTh1} & \multicolumn{1}{c|}{{\color[HTML]{FF0000} \textbf{0.655 0.567}}} & \multicolumn{1}{c|}{0.766 0.611} & \multicolumn{1}{c|}{0.745 0.610} & \multicolumn{1}{c|}{0.840 0.631} & \multicolumn{1}{c|}{0.734 0.833} & \multicolumn{1}{c|}{0.699 {\color[HTML]{00B0F0} {\ul 0.588}}} & \multicolumn{1}{c|}{{\color[HTML]{00B0F0} {\ul 0.696}} 0.624} & \multicolumn{1}{c|}{0.921 1.091} & \multicolumn{1}{c|}{0.765 0.637} & \multicolumn{1}{c|}{1.083 0.831} & 0.808 0.675 \\ \midrule
    \multicolumn{1}{c|}{ETTh2} & \multicolumn{1}{c|}{{\color[HTML]{FF0000} \textbf{0.480 0.480}}} & \multicolumn{1}{c|}{0.541 0.518} & \multicolumn{1}{c|}{0.513 \color[HTML]{00B0F0} {\ul 0.495}} & \multicolumn{1}{c|}{0.563 0.523} & \multicolumn{1}{c|}{0.547 0.537} & \multicolumn{1}{c|}{{\color[HTML]{00B0F0} {\ul 0.508}} 0.498} & \multicolumn{1}{c|}{1.218 0.787} & \multicolumn{1}{c|}{2.530 1.233} & \multicolumn{1}{c|}{0.625 0.574} & \multicolumn{1}{c|}{3.263 1.509} & 0.658 0.648 \\ \midrule
    \multicolumn{1}{c|}{ETTm1} & \multicolumn{1}{c|}{{\color[HTML]{FF0000} \textbf{0.484}} \color[HTML]{00B0F0} {\ul 0.463}} & \multicolumn{1}{c|}{0.554 0.495} & \multicolumn{1}{c|}{0.544 0.480} & \multicolumn{1}{c|}{0.523 0.483} & \multicolumn{1}{c|}{0.532 0.476} & \multicolumn{1}{c|}{{\color[HTML]{00B0F0} {\ul 0.503}} {\color[HTML]{FF0000} \textbf{0.460}}} & \multicolumn{1}{c|}{0.540 0.498} & \multicolumn{1}{c|}{3.555 1.483} & \multicolumn{1}{c|}{0.522 0.501} & \multicolumn{1}{c|}{1.093 0.811} & 0.631 0.550 \\ \midrule
    \multicolumn{1}{c|}{ETTm2} & \multicolumn{1}{c|}{{\color[HTML]{FF0000} \textbf{0.425 0.434}}} & \multicolumn{1}{c|}{0.468 0.449} & \multicolumn{1}{c|}{0.464 \color[HTML]{00B0F0}{\ul 0.447}} & \multicolumn{1}{c|}{0.465 0.449} & \multicolumn{1}{c|}{{\color[HTML]{00B0F0} {\ul 0.446}} {\color[HTML]{FF0000} \textbf{0.434}}} & \multicolumn{1}{c|}{0.462 0.448} & \multicolumn{1}{c|}{0.655 0.574} & \multicolumn{1}{c|}{3.555 1.483} & \multicolumn{1}{c|}{0.487 0.475} & \multicolumn{1}{c|}{4.566 1.745} & 0.516 0.491 \\
    
    \bottomrule
    \end{tabular}}
\end{table}

\subsection{Ablation  Studies}\label{ablation studies}
\textbf{AHL Module.} To investigate the effectiveness of the AHL model, we conduct ablation studies by designing the following three variations: (1) Replacing the AHL module with adaptive graph learning module (-AGL). (2) Replacing the AHL model with one incidence matrix to capture group-wise node interactions at different scales (-one). (3) Replacing the AHL module with predefined multi-scale hypergraphs (-PH), i.e., each hyperedge connected a fixed number of nodes (4 in our experiment) in chronological order. The experimental results on ETTh1 dataset are shown in Table \ref{tab:table5}. We can observe that -AGL gets the worst forecasting results, indicating the importance of modeling group-wise interactions. In addition, -PH and -one perform worse than Ada-MSHyper, showing the effectiveness of adaptive hypergraph and multi-scale hypergraph, respectively.
\begin{table}[htbp]
\centering
\caption{Results of different adaptive hypergraph learning methods and constraint mechanisms.}
\label{tab:table5}
\resizebox{0.8 \linewidth}{!}{
\begin{tabular}{@{}c|cccccc|cccccc|cc@{}}
\toprule
Variation & \multicolumn{2}{c}{AGL} & \multicolumn{2}{c}{one} & \multicolumn{2}{c|}{PH} & \multicolumn{2}{c}{w/o NC} & \multicolumn{2}{c}{w/o HC} & \multicolumn{2}{c|}{w/o NHC} & \multicolumn{2}{c}{Ada-MSHyper} \\ \midrule
Metric & MSE & MAE & MSE & MAE & MSE & MAE & MSE & MAE & MSE & MAE & MSE & MAE & MSE & MAE \\ \midrule
96 & 0.542 & 0.560 & 0.422 & 0.437 & 0.386 & 0.403 & 0.390 & 0.403 & 0.384 & 0.416 & 0.393 & 0.422 & \textbf{0.372} & \textbf{0.393} \\
336 & -- & -- & 0.559 & 0.502 & 0.448 & 0.452 & 0.423 & 0.437 & 0.430 & 0.435 & 0.425 & 0.441 & \textbf{0.422} & \textbf{0.433} \\
720 & -- & -- & 0.563 & 0.617 & 0.456 & 0.458 & 0.448 & \textbf{0.457} & 0.451 & 0.460 & 0.449 & 0.466 & \textbf{0.445} & {0.459} \\ \bottomrule
\end{tabular}}
\end{table}

\textbf{NHC Mechanism.} To investigate the effectiveness of the NHC mechanism, we conduct ablation studies by designing the following three variations: (1) Removing the node constraint (-w/o NC). (2) Removing the hyperedge constraint (-w/o HC). (3) Removing the NHC mechanism (-w/o NHC). The experimental results on ETTh1 dataset are shown in Table \ref{tab:table5}. We can observe that Ada-MSHyper performs better than -w/o NC and -w/o HC, showing the effectiveness of node constraint and hyperedge constraint, respectively. In addition, -w/o NHC gets the worst forecasting results, which demonstrates the superiority of the NHC mechanism in adaptive hypergraph learning. More results about ablation studies are shown in Appendix \ref{Ablation Studies2}. 

\begin{figure*}[htbp]
\centering
\subfloat[Input sequence]{\includegraphics[width=0.24\textwidth]{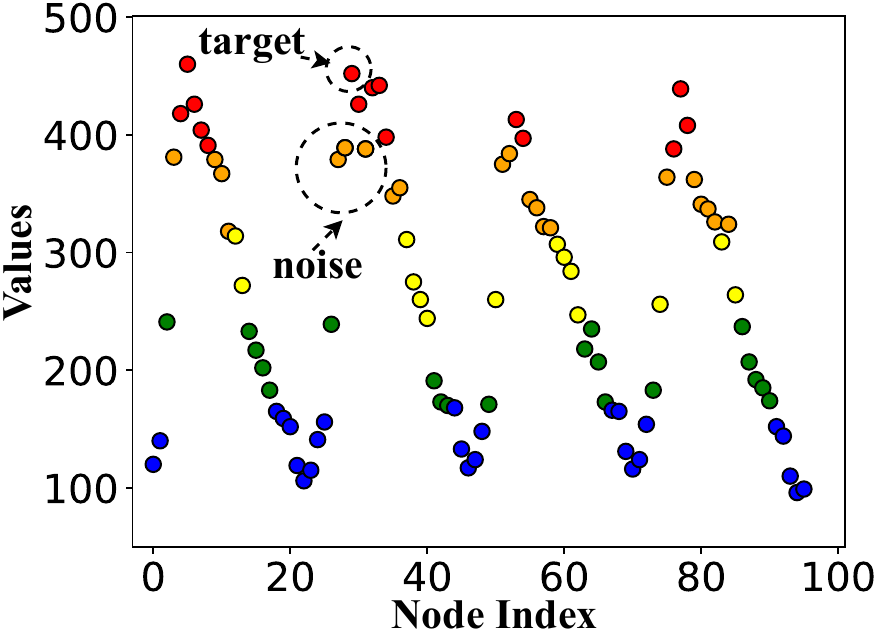}%
\label{fig_first_case}}
\vspace{-1pt}
\subfloat[-w/o NHC]{\includegraphics[width=0.24\textwidth]{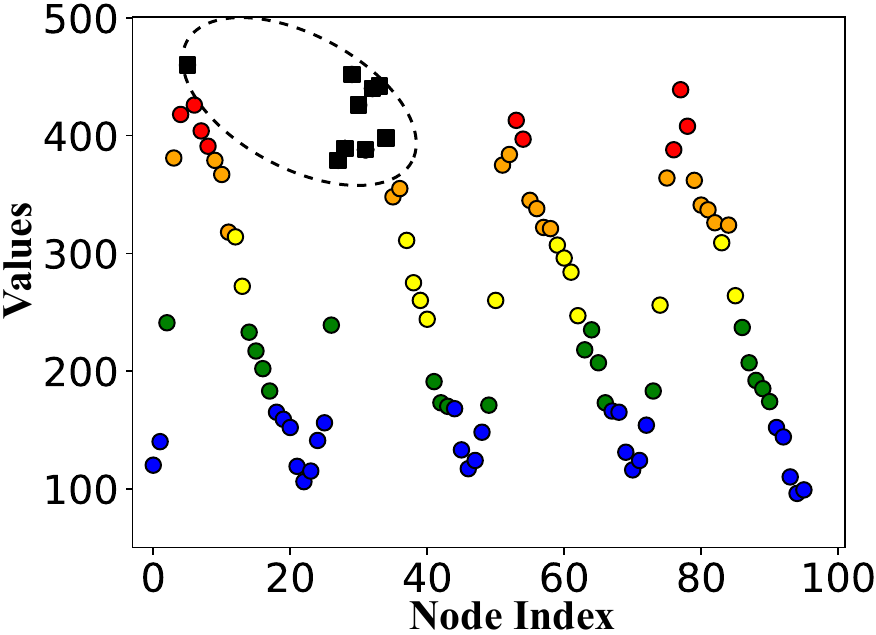}%
\label{fig_fourth_case}}
\vspace{-1pt}
\subfloat[-w/o HC]{\includegraphics[width=0.24\textwidth]{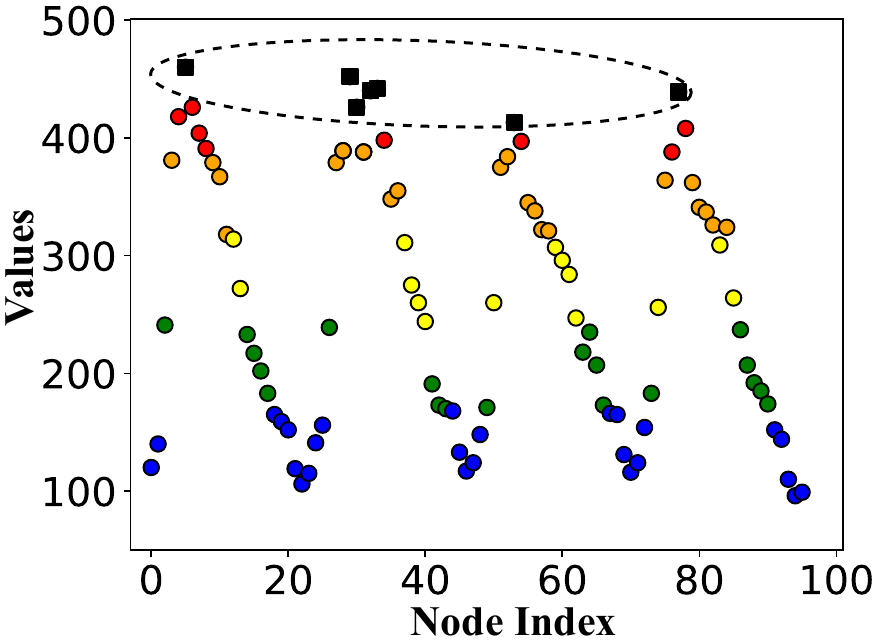}%
\label{fig_second_case}}
\vspace{-1pt}
\subfloat[Ada-MSHyper]{\includegraphics[width=0.24\textwidth]{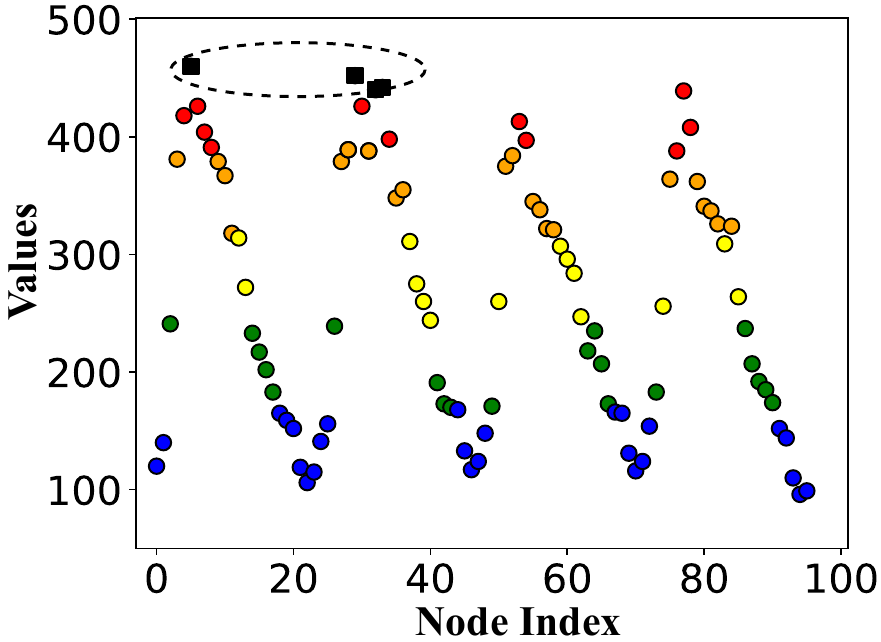}%
\label{fig_third_case}}
\vspace{-1pt}
\caption{Visualization the node constraint effect on Electricity dataset.}
\label{Figure_4}
\end{figure*}

To further demonstrate the effectiveness of the node constraint in clustering nodes with similar semantic information, we present case visualization with -w/o NHC and -w/o HC on Electricity dataset. We randomly select one sample and plot the node values at the finest scale. We categorize the nodes into four groups based on the node values. Nodes with the same color indicate that they may have similar semantic information. As shown in Figure \ref{fig_first_case}, for the target node, nodes of other colors may be considered as noise. We drew the nodes related to the target node in black color based on incidence matrix $\mathbf{H}^1$. As shown in Figure \ref{fig_fourth_case}, due to the lack of node constraint, -w/o NHC can only capture the interactions among the target node and neighboring nodes and cannot distinguish nuanced noise information. In Figure \ref{fig_second_case}, with the node constraint, -w/o HC can cluster neighboring and distant but still strongly correlated nodes. In Figure \ref{fig_third_case}, with the NHC mechanism, Ada-MSHyper cannot only cluster nodes with similar semantic information but can differentiate temporal variations. The full visualization results are shown in Appendix \ref{case appedix}.

\begin{wrapfigure}{r}{0.5\textwidth}
\includegraphics[width=2.9in]{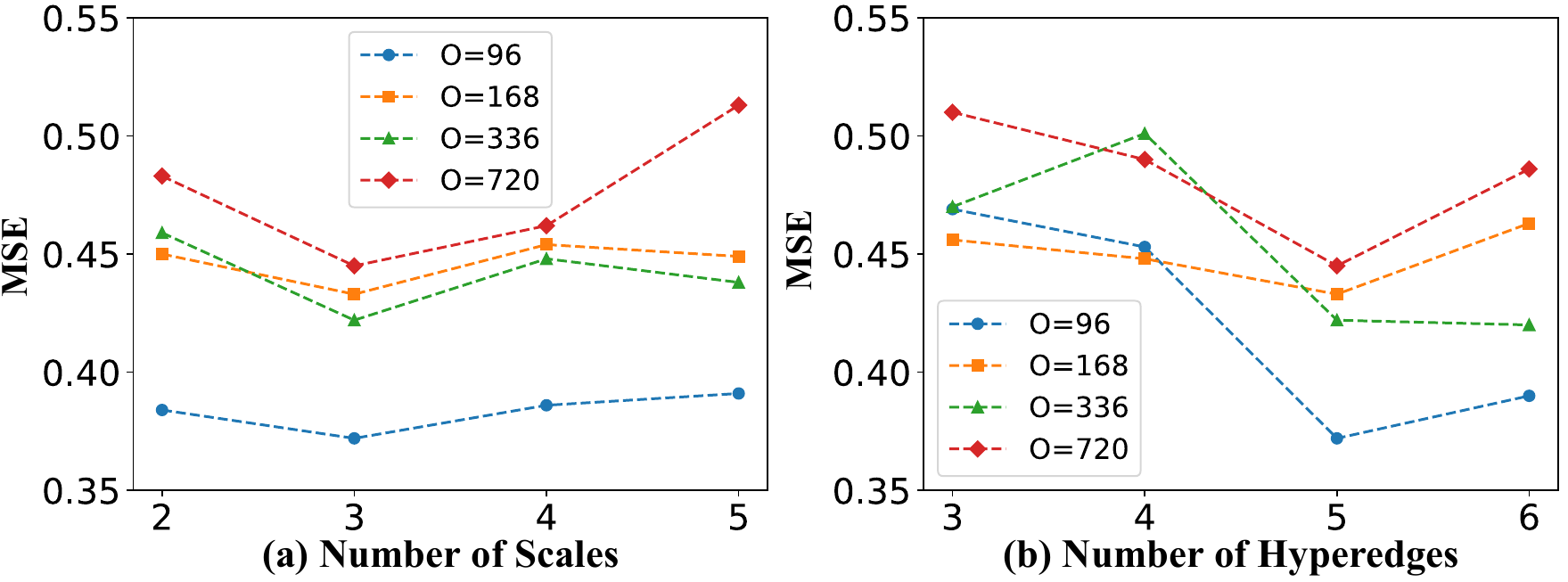}
\centering
\caption{The impact of hyperparameters.}
\label{Figure_5}
\end{wrapfigure}
\subsection{Parameter Studies}
We perform parameter studies to measure the impact of the number of scales (\#scales) and the max number of hyperedges connected to a node (\#hyperedges). The experimental results on ETTh1 dataset are shown in Figure \ref{Figure_5}, we can see that: (1) the best performance can be obtained when \#scales is 3. The reason is that smaller \#scales cannot provide sufficient pattern information and larger \#scales may introduce excessive parameters and result in overfitting problems. (2) The optimal \#hyperedges is 5. The reason is that smaller \#hyperedges cannot capture group-wise interactions sufficiently and larger \#hyperedges may introduce noise. More results about parameter studies are shown in Appendix \ref{Parameter Studies}. 

\subsection{Computational Cost}
We compare Ada-MSHyper with the two latest transformer-based methods, i.e., iTransformer and PatchTST, on traffic datasets with the output length of 96. The experimental results are shown in Table \ref{tab:computationcost}. Although we have a larger number of parameters, we achieve lower training time and lower GPU occupation due to the matrix sparsity operation in the model and the optimization of hypergraph computation provided by \textit{torch\_geometry}\cite{HAHC}. Considering the forecasting performance and the computation cost, Ada-MSHyper demonstrates its superiority over existing methods. 
\begin{table}[!h]
\centering
\begin{tiny}
  \caption{Computation cost.}
  \label{tab:computationcost}
  \resizebox{0.6 \textwidth}{!}{
    \begin{tabular}{l|c|c|c|c}
    \toprule
    Methods & Training Time & \# Parameters & GPU Occupation & MSE results \\
    \midrule
    Ada-MSHyper & \textbf{6.499s} & 8,965,392 & \textbf{6,542MB} & \textbf{0.384} \\
    iTransformer & 7.863s & 6,731,984 & 6,738MB & 0.395 \\
    PatchTST & 17.603s & \textbf{548,704} & 9,788MB & 0.526 \\
    \bottomrule
    \end{tabular}%
    }
\end{tiny}
\end{table}
\section{Conclusions and Future Work}
In this paper, we propose Ada-MSHyper with an adaptive multi-scale hypergraph for time series forecasting. Empowered by the AHL module and multi-scale interaction module, Ada-MSHyper can promote more comprehensive multi-scale group-wise pattern interactions, addressing the problem of semantic information sparsity. Experimentally, Ada-MSHyper achieves the SOTA performance, reducing prediction errors by an average of 4.56\%, 10.38\%, and 4.97\% in MSE for long-range, short-range, and ultra-long-range time series forecasting, respectively. In addition, the visualization analysis and the ablation studies demonstrate the effectiveness of NHC mechanism in clustering nodes with similar semantic information and in addressing the issue of temporal variations entanglement. 

In the future, this work can be extended in the following two aspects. First, since 2D spectrogram data may offer a better representation for time series forecasting, we will adapt our framework to the 2D spectrogram data in time-frequency domain. Second, since the features extracted by the MFE module may contain redundant information, we will design a disentangled multi-scale feature extraction module to extract more independent and representative features. 

\section{Acknowledgement}
This work was supported by the Science Foundation of Donghai Laboratory (Grant No. DH-2022ZY0013).

\bibliographystyle{plain}
\bibliography{ref1}

\newpage
\appendix

\section{Descriptions of Notations}\label{notations}
To help understand the symbols used throughout the paper, we provide a detailed list of the key notations in Table \ref{tab:nota}.

\begin{table}[htbp]
    \centering
    \begin{tiny}
    \caption{Description of the key notations.}
    \begin{tabular}{l|m{5.5cm}}
        \toprule
        Notation & Descriptions  \\ \midrule
        $\mathcal{G}$ & Hypergraph\\ \midrule
        $\mathcal{E}$& Hyperedge set \\ \midrule
        $\mathcal{V}$& Node set \\ \midrule
        $N$ & Number of nodes \\ \midrule 
        $M$ & Number of hyperedges \\ \midrule 
        $T$ & Input length \\ \midrule 
        $H$ & Output length \\ \midrule 
        $D$ & Feature dimension\\ \midrule 
        $S$ & Total number of temporal scales \\ \midrule 
         $s$ & Scale index\\ \midrule 
         $\mathbf{X}_{1: T}^{\text{I}}$ & Historical input sequence \\ \midrule 
         $\mathbf{X}^s$ & Sub-sequence at scale $s$  \\ \midrule
        $\boldsymbol{x}_t $ & Values at time step $t$\\ \midrule
        
        $\widehat{\mathbf X}_{T+1:T+H}^\text{O} \in\mathbb{R}^{H \times D}$ & Forecasting results\\ \midrule 
        $\boldsymbol{E}_{\mathrm{node}}^s\in \mathbb{R}^{N^s\times D}$ & Node embeddings at scale $s$ \\ \midrule
        $\boldsymbol{E}_{\mathrm{hyper}}^{s}\in \mathbb{R}^{M^s\times D}$  & Hyperedge embeddings at scale $s$ \\ \midrule 
         $e_i^s$ & $i$th hyperedge at scale $s$ \\ \midrule 
         $v_i^s$ & $i$th node at scale $s$ \\ \midrule
         $\boldsymbol{e}_{i}^s$ & $i$th hyperedge feature representation at scale $s$ \\ \midrule
          $\boldsymbol{v}_{i}^s$ & $i$th node feature representation at scale $s$ \\ \midrule
        $\eta$ & Threshold of $TopK$ function \\ \midrule
        $\beta$ & Threshold of the scale-specific incidence matrices\\ \midrule
        $\gamma$ & Threshold of the Euclidean distance\\ \midrule
        $\lambda$ &  Balancing hyperparameter between node loss and hyperedge loss\\ \midrule
        $\mathbf{H}^s$ & Incidence matrix at scale $s$\\ \midrule
        $l^{s-1}$ & Size of the aggregation window at scale $s-1$\\ \midrule
        $\boldsymbol{\mathcal{V}}^s$ & Initialized node feature representations at scale $s$\\  \midrule
        $\boldsymbol{\mathcal{E}}^s$ & Initialized hyperedge feature representations at scale $s$\\  \midrule
         $\boldsymbol{\widetilde{\mathcal{V}}}^s$ & Updated node feature representations at scale $s$ \\ \midrule
        $\tilde{\mathbf{V}}$ & Updated hyperedge feature representations \\ \midrule
        $\mathcal{J}$ & Number of heads \\ \midrule
        $\boldsymbol{\mathcal{H}}^{s}_{\jmath}$ & Enriched incidence matrix of the $\jmath$th head at scale $s$ \\ \midrule
        $\mathcal{N}(v_i^s)$ & Neighboring hyperedges connected to $v_i^s$ \\ \midrule
      $\mathcal{N}(e_i^s)$ & Neighboring nodes connected by $e_i^s$\\ \midrule
          $\mathbf{Q}$,  $\mathbf{K}$,  $\mathbf{V}\in\mathbb{R}^{M\times D}$ & Queries, keys, and values \\ \midrule
          $\alpha_{i,j}$ & Correlation weight between $i$th and $j$th hyperedge representations\\ \midrule
          $D_{i,j}$ & Euclidean distance between $i$th and $j$th hyperedge representations \\ \midrule
        $L_{node}^s$ & Node constraint loss at scale $s$ \\ \midrule
        $L_{hyper}^s$ & Hyperedges constraint loss at scale $s$ \\ \midrule
        $L_{const}$ & Final constraint loss \\ \midrule
        $L_{mse}$ & MSE loss \\ \bottomrule       
    \end{tabular}
    \label{tab:nota}
    \end{tiny}
\end{table}

\section{Descriptions of Datasets} \label{datasets}
\textbf{Datasets.} For long-range time series forecasting, we conduct experiments on 7 commonly used benchmarks, including Electricity Transformers Temperature (ETT), Traffic\footnote{\url {http://pems.dot.ca.gov}}, Electricity\footnote{\url{https://archive.ics.uci.edu/ml/datasets/ElectricityLoadDiagrams20112014}}, and Weather\footnote{
\url { https://www.bgc-jena.mpg.de/wetter/}} datasets following \cite{Autoformer, Itransformer, MSHyper}. 
ETT datasets include data from two counties in the same Chinese province, each data point comprising seven variables: the target variable "oil temperature" and six power load features. The datasets vary in granularity, with "h" indicating hourly data and "m" indicating 15-minute intervals.
Weather dataset contains 21 weather indicators collected every 10 minutes from a weather station in Germany.
Electricity dataset records hourly electricity consumption of 321 clients.
Traffic dataset provides hourly road occupancy rates from 821 freeway sensors.
For short-range time series forecasting, we use four benchmarks from PEMS (PEMS03, PEMS04, PEMS07, and PEMS08), as referenced in \cite{timemixer, scinet}. These datasets capture 5-minute traffic flow data from freeway sensors. For ultra-long-range time series forecasting, we adopt ETTh1, ETTh2,  ETTm1, and ETTm2 following \cite{witran}. Table \ref{tab:addtable11} gives the detailed dataset statistics. In addition, the forecastability is derived from one minus the entropy of the Fourier decomposition of a time series\cite{timemixer, forecastable}. Higher values mean greater forecastability.

\begin{table}[htbp]
\centering
  \begin{tiny}
      \caption{Detailed dataset statistics.}
      \label{tab:addtable11}
      \begin{tabular}{c|l|l|l|l|l|l}
        \toprule
        Task & Dataset & \# Variates & Prediction Length & Frequency & Forecastability & Information \\
        \midrule
        \multirow{5}[1]{*}{Long-term} & ETTh1, ETTh2 & 7 & (96, 192, 336,720) & Hourly & 0.38, 0.45 & Temperature \\
        \cmidrule{2-7}
        & ETTm1, ETTm2 & 7 & (96, 192, 336,720) & 15 mins & 0.46, 0.55 & Temperature \\
        \cmidrule{2-7}
        & Weather & 21 & (96, 192, 336,720) & 10 mins & 0.75 & Weather \\
        \cmidrule{2-7}
        & Electricity & 321 & (96, 192, 336,720) & Hourly & 0.77 & Electricity \\
        \cmidrule{2-7}
        & Traffic & 862 & (96, 192, 336,720) & Hourly & 0.68 & Transportation \\
        \midrule
        \multirow{4}[1]{*}{Short-term} & PEMS03 & 358 & 12 & 5min & 0.65 & Transportation \\
        \cmidrule{2-7}
        & PEMS04 & 307 & 12 & 5mins & 0.45 & Transportation \\
        \cmidrule{2-7}
        & PEMS07 & 883 & 12 & 5mins & 0.58 & Transportation \\
        \cmidrule{2-7}
        & PEMS08 & 170 & 12 & 5mins & 0.52 & Transportation \\
        \bottomrule
      \end{tabular}
  \end{tiny}
\end{table}
We adopt the same data processing and train-validation-test split protocol as in existing works \cite{MSHyper, Itransformer, PatchTST}. We split each dataset into training, validation, and test sets based on chronological order. For PEMS (PEMS03, PEMS04, PEMS07, and PEMS08) dataset and ETT (ETTh1, ETTh2, ETTm1, and ETTm2) dataset, the train-validation-test split ratio is 6:2:2. For Weather, Traffic, and Electricity dataset, the train-validation-test split ratio is 7:2:1.

\textbf{Metric details.} Following existing methods \cite{MSHyper, Itransformer}, we employ Mean Squared Error (MSE) and Mean Absolute Error (MAE) as our evaluation metrics, which can be formulated as follows:
\begin{equation}
\label{26}
 L_{mse}=\frac{1}{H}\left\|\widehat{\mathbf X}_{T+1:T+H}^\text{O}-\mathbf X_{T+1:T+H}^\text{O}\right\|_2^2
\end{equation}
\begin{equation}
\label{27}
 L_{mae}=\frac{1}{H}\big|\widehat{\mathbf X}_{T+1:T+H}^\text{O}-\mathbf X_{T+1:T+H}^\text{O}\big|,
\end{equation}
where $T$ and $H$ are the input and output lengths, $\widehat{\mathbf X}_{T+1:T+H}^\text{O}$ and $\mathbf X_{T+1:T+H}^\text{O}$ are the predicted results and ground truth.

\section{Descriptions of Baselines}\label{baselines}
We compare Ada-MSHyper with 15 competitive baselines. Below are brief descriptions of the baselines:
(1) iTransformer \cite{Itransformer}: Applies the attention and feed-forward network on the inverted dimensions, i.e., the time points of individual series are embedded into variate tokens, and the feed-forward network is applied for each variate token to learn nonlinear representations. 
(2) MSHyper \cite{MSHyper}: Utilizes rule-based multi-scale hypergraphs to model high-order pattern interactions in univariate time series. 
(3) PatchTST \cite{PatchTST}: Uses channel-independent techniques and treats subseries-level patches as input tokens to a Transformer, facilitating semantic extraction of multiple time steps in time series. 
(4) TimesMixer \cite{timemixer}: Employs a fully MLP-based architecture with past-decomposable-mixing and future-multipredictor-mixing blocks to leverage disentangled multiscale series. 
(5) MSGNet \cite{MSGNet}: Leverages frequency domain analysis to extract periodic patterns and combines an attention mechanism with adaptive graph convolution to capture multi-scale pattern interactions.
(6) CrossGNN \cite{CrossGNN}: Uses an adaptive multi-scale identifier to construct multi-scale representations and utilizes a cross-scale GNN to capture multi-scale pattern interactions.
(7) TimesNet \cite{timesnet}: Conducts multi-periodicity analysis by extending 1D time series into a set of 2D tensors, modeling complex temporal variations from a 2D perspective. 
(8) WITRAN \cite{witran}: Proposes an RNN-based architecture that handles univariate input sequences from a 2D space perspective, maintaining a fixed scale throughout the processing. 
(9) SCINet \cite{scinet}: Uses a recursive downsample-convolve-interact architecture to extract temporal features from downsampled sub-sequences or features. 
(10) Crossformer \cite{Crossformer}: Adopts cross-dimension attention to capture inter-series dependencies for multivariate time series forecasting.
(11) FiLM \cite{film}: Applies Legendre polynomial projections to approximate historical information, uses Fourier projections to remove noise, and adds a low-rank approximation to speed up computation.
(12) DLinear \cite{Dlinear}: Decomposes time series into two different components and uses a single linear layer for each component to model temporal dependencies.
(13) FEDformer \cite{FEDformer}: Utilizes a seasonal-trend decomposition method to capture the global profile of time series and a frequency-enhanced Transformer to capture more detailed structures.
(14) Pyraformer \cite{pyraformer}: Utilizes a pyramidal attention module to extract inter-scale features at different resolutions and intra-scale features at different ranges with linear complexity.
(15) Autoformer \cite{Autoformer}: Uses an auto-correlation mechanism based on series periodicity to capture features at the sub-series level.

\section{Experimental Settings}\label{experimental settings}
We repeat all experiments 3 times and use the mean of the metrics as the final results. The training process is early stopped when there is no improvement within 5 epochs. Following existing works \cite{Itransformer, timemixer, Dlinear}, we use instance normalization to normalize all datasets. The max number of scale $S$ is set to 3. 
We use 1D convolution as our aggregation function. For other hyperparameters, we use Neural Network Intelligence (NNI)\footnote{\url{https://nni.readthedocs.io/en/latest/}} toolkit to automatically search the best hyperparameters, which can greatly reduce computation cost compared to the grid search approach. The detailed search space of hyperparameters is given in Table \ref{tab:hyper}. The source code of Ada-MSHyper is released on GitHub \footnote{\url{https://github.com/shangzongjiang/Ada-MSHyper}}.

\begin{table}[htbp]
    \centering
\begin{tiny}
    \caption{The search space of hyperparameters.}
    \resizebox{0.4\linewidth}{!}{
    \begin{tabular}{l|l}
        \toprule
        {Parameters} & {Choise}  \\ \midrule
        Batch size & \{8, 16, 32, 64, 128\}\\ \midrule
        Number of hyperedges at scale 1 & \{10, 20, 30, 50\} \\ \midrule
        Number of hyperedges at scale 2 & \{5, 10, 15, 20\}\\ \midrule
        Number of hyperedges at scale 3 & \{1, 2, 4, 5, 8, 12\}\\ \midrule
        Aggregation window at scale 1 & \{2, 4, 8\}\\ \midrule
        Aggregation window at scale 2 & \{2, 4\}\\ \midrule
        $\eta$ & \{1, 3, 5, 10, 15, 20\}\\ \midrule
        $\beta$ & \{0.2, 0.3, 0.4, 0.5\}\\ \midrule
        $\gamma$ & \{0.2, 0.3, 0.4, 0.5\} \\ \bottomrule
    \end{tabular}
    }
    \label{tab:hyper}
    \end{tiny}
\end{table}

\section{Full Results} \label{full_res}
We compare Ada-MSHyper with 13 baselines across four tasks: long-range forecasting for multivariate time series, long-range forecasting for univariate time series, ultra-long-range forecasting for multivariate time series, and short-range forecasting for multivariate time series. For a fair comparison, we evaluate Ada-MSHyper and baselines under unified experimental settings of each task. 
The average results from all prediction lengths are presented in tables, where the best results are \textbf{\textcolor{red}{bolded}} and the second best results are {\color[HTML]{00CCFF} {\ul underlined}}. * indicates that some baselines do not meet our settings, thus we rerun these baselines using their official code and fine-tune their key hyperparameters. 

\textbf{Long-Range Time Series Forecasting Under Multivariate Settings.} Table \ref{tab:addlabel1} summarizes the results of long-range time series forecasting under multivariate settings, where the results of baselines without * are cited from iTransformer \cite{Itransformer}. We can see from Table \ref{tab:addlabel1} that Ada-MSHyper achieves the SOTA results on all datasets. Specifically, Ada-MSHyper gives an average error reduction of 4.56\% and 3.47\% compared to the best baseline in MSE and MAE, respectively.

\begin{table}[htbp]
\caption{Full results of long-range time series forecasting under multivariate settings.}
\resizebox{\linewidth}{!}{
\begin{tabular}{cccccccccccccccccccccccccccc}
\toprule
\multicolumn{2}{c}{Models} & \multicolumn{2}{c}{\begin{tabular}[c]{@{}c@{}}Ada-MSHyper\\ (Ours)\end{tabular}} & \multicolumn{2}{c}{\begin{tabular}[c]{@{}c@{}}iTransformer\\ (2024)\end{tabular}} & \multicolumn{2}{c}{\begin{tabular}[c]{@{}c@{}}MSHyper*\\      (2024)\end{tabular}} & \multicolumn{2}{c}{\begin{tabular}[c]{@{}c@{}}TimeMixer*\\ (2024)\end{tabular}} & \multicolumn{2}{c}{\begin{tabular}[c]{@{}c@{}}MSGNet*\\ (2024)\end{tabular}} & \multicolumn{2}{c}{\begin{tabular}[c]{@{}c@{}}CrossGNN*\\  (2023)\end{tabular}} & \multicolumn{2}{c}{\begin{tabular}[c]{@{}c@{}}PatchTST\\      (2023)\end{tabular}} & \multicolumn{2}{c}{\begin{tabular}[c]{@{}c@{}}Crossformer\\      (2023)\end{tabular}} & \multicolumn{2}{c}{\begin{tabular}[c]{@{}c@{}}TimesNet\\      (2023)\end{tabular}} & \multicolumn{2}{c}{\begin{tabular}[c]{@{}c@{}}DLinear\\      (2023)\end{tabular}} & \multicolumn{2}{c}{\begin{tabular}[c]{@{}c@{}}FiLM*\\      (2022)\end{tabular}} & \multicolumn{2}{c}{\begin{tabular}[c]{@{}c@{}}FEDformer\\      (2022)\end{tabular}} & \multicolumn{2}{c}{\begin{tabular}[c]{@{}c@{}}Autoformer\\      (2021)\end{tabular}} \\ \midrule
\multicolumn{2}{c|}{Metric} & MSE & \multicolumn{1}{c|}{MAE} & MSE & \multicolumn{1}{c|}{MAE} & MSE & \multicolumn{1}{c|}{MAE} & MSE & \multicolumn{1}{c|}{MAE} & MSE & \multicolumn{1}{c|}{MAE} & MSE & \multicolumn{1}{c|}{MAE} & MSE & \multicolumn{1}{c|}{MAE} & MSE & \multicolumn{1}{c|}{MAE} & MSE & \multicolumn{1}{c|}{MAE} & MSE & \multicolumn{1}{c|}{MAE} & MSE & \multicolumn{1}{c|}{MAE} & MSE & \multicolumn{1}{c|}{MAE} & MSE & MAE \\ \midrule
\multicolumn{1}{c|}{} & \multicolumn{1}{c|}{96} & {\color[HTML]{FF0000} \textbf{0.157}} & \multicolumn{1}{c|}{{\color[HTML]{FF0000} \textbf{0.195}}} & 0.174 & \multicolumn{1}{c|}{0.214} & 0.170 & \multicolumn{1}{c|}{0.223} & 0.163 & \multicolumn{1}{c|}{{\color[HTML]{00B0F0} {\ul 0.210}}} & 0.163 & \multicolumn{1}{c|}{0.212} & 0.159 & \multicolumn{1}{c|}{0.218} & 0.177 & \multicolumn{1}{c|}{0.218} & {\color[HTML]{00B0F0} {\ul 0.158}} & \multicolumn{1}{c|}{0.230} & 0.172 & \multicolumn{1}{c|}{0.220} & 0.196 & \multicolumn{1}{c|}{0.255} & 0.199 & \multicolumn{1}{c|}{0.262} & 0.217 & \multicolumn{1}{c|}{0.296} & 0.266 & 0.336 \\
\multicolumn{1}{c|}{} & \multicolumn{1}{c|}{192} & 0.218 & \multicolumn{1}{c|}{0.259} & 0.221 & \multicolumn{1}{c|}{{\color[HTML]{00B0F0} {\ul 0.254}}} & 0.218 & \multicolumn{1}{c|}{{\color[HTML]{FF0000} \textbf{0.253}}} & 0.212 & \multicolumn{1}{c|}{0.257} & 0.212 & \multicolumn{1}{c|}{{\color[HTML]{00B0F0} {\ul 0.254}}} & {\color[HTML]{00B0F0} {\ul 0.211}} & \multicolumn{1}{c|}{0.266} & 0.225 & \multicolumn{1}{c|}{0.259} & {\color[HTML]{FF0000} \textbf{0.206}} & \multicolumn{1}{c|}{0.277} & 0.219 & \multicolumn{1}{c|}{0.261} & 0.237 & \multicolumn{1}{c|}{0.296} & 0.228 & \multicolumn{1}{c|}{0.288} & 0.276 & \multicolumn{1}{c|}{0.336} & 0.307 & 0.367 \\
\multicolumn{1}{c|}{} & \multicolumn{1}{c|}{336} & {\color[HTML]{FF0000} \textbf{0.251}} & \multicolumn{1}{c|}{{\color[HTML]{FF0000} \textbf{0.252}}} & 0.278 & \multicolumn{1}{c|}{0.296} & 0.269 & \multicolumn{1}{c|}{0.300} & {\color[HTML]{00B0F0} {\ul 0.263}} & \multicolumn{1}{c|}{{\color[HTML]{00B0F0} {\ul 0.292}}} & 0.272 & \multicolumn{1}{c|}{0.299} & 0.267 & \multicolumn{1}{c|}{0.310} & 0.278 & \multicolumn{1}{c|}{0.297} & 0.272 & \multicolumn{1}{c|}{0.335} & 0.280 & \multicolumn{1}{c|}{0.306} & 0.283 & \multicolumn{1}{c|}{0.335} & 0.267 & \multicolumn{1}{c|}{0.323} & 0.339 & \multicolumn{1}{c|}{0.380} & 0.359 & 0.395 \\
\multicolumn{1}{c|}{\multirow{-4}{*}{Weather}} & \multicolumn{1}{c|}{720} & {\color[HTML]{FF0000} \textbf{0.304}} & \multicolumn{1}{c|}{{\color[HTML]{FF0000} \textbf{0.328}}} & 0.358 & \multicolumn{1}{c|}{0.347} & 0.343 & \multicolumn{1}{c|}{{\color[HTML]{00B0F0} {\ul 0.341}}} & 0.343 & \multicolumn{1}{c|}{0.345} & 0.350 & \multicolumn{1}{c|}{0.348} & 0.352 & \multicolumn{1}{c|}{0.362} & 0.354 & \multicolumn{1}{c|}{0.348} & 0.398 & \multicolumn{1}{c|}{0.418} & 0.365 & \multicolumn{1}{c|}{0.359} & 0.345 & \multicolumn{1}{c|}{0.381} & {\color[HTML]{00B0F0} {\ul 0.319}} & \multicolumn{1}{c|}{0.361} & 0.403 & \multicolumn{1}{c|}{0.428} & 0.419 & 0.428 \\ \midrule
\multicolumn{1}{c|}{} & \multicolumn{1}{c|}{96} & {\color[HTML]{FF0000} \textbf{0.135}} & \multicolumn{1}{c|}{{\color[HTML]{FF0000} \textbf{0.238}}} & {\color[HTML]{00B0F0} {\ul 0.148}} & \multicolumn{1}{c|}{{\color[HTML]{00B0F0} {\ul 0.240}}} & 0.176 & \multicolumn{1}{c|}{0.261} & 0.153 & \multicolumn{1}{c|}{0.247} & 0.165 & \multicolumn{1}{c|}{0.274} & 0.173 & \multicolumn{1}{c|}{0.275} & 0.181 & \multicolumn{1}{c|}{0.270} & 0.219 & \multicolumn{1}{c|}{0.314} & 0.168 & \multicolumn{1}{c|}{0.272} & 0.197 & \multicolumn{1}{c|}{0.282} & 0.198 & \multicolumn{1}{c|}{0.274} & 0.193 & \multicolumn{1}{c|}{0.308} & 0.201 & 0.317 \\
\multicolumn{1}{c|}{} & \multicolumn{1}{c|}{192} & {\color[HTML]{FF0000} \textbf{0.152}} & \multicolumn{1}{c|}{{\color[HTML]{FF0000} \textbf{0.239}}} & {\color[HTML]{00B0F0} {\ul 0.162}} & \multicolumn{1}{c|}{{\color[HTML]{00B0F0} {\ul 0.253}}} & 0.173 & \multicolumn{1}{c|}{0.260} & 0.166 & \multicolumn{1}{c|}{0.256} & 0.184 & \multicolumn{1}{c|}{0.292} & 0.195 & \multicolumn{1}{c|}{0.288} & 0.188 & \multicolumn{1}{c|}{0.274} & 0.231 & \multicolumn{1}{c|}{0.322} & 0.184 & \multicolumn{1}{c|}{0.289} & 0.196 & \multicolumn{1}{c|}{0.285} & 0.198 & \multicolumn{1}{c|}{0.278} & 0.201 & \multicolumn{1}{c|}{0.315} & 0.222 & 0.334 \\
\multicolumn{1}{c|}{} & \multicolumn{1}{c|}{336} & {\color[HTML]{FF0000} \textbf{0.168}} & \multicolumn{1}{c|}{{\color[HTML]{FF0000} \textbf{0.266}}} & {\color[HTML]{00B0F0} {\ul 0.178}} & \multicolumn{1}{c|}{{\color[HTML]{00B0F0} {\ul 0.269}}} & 0.195 & \multicolumn{1}{c|}{0.297} & 0.185 & \multicolumn{1}{c|}{0.277} & 0.195 & \multicolumn{1}{c|}{0.302} & 0.206 & \multicolumn{1}{c|}{0.300} & 0.204 & \multicolumn{1}{c|}{0.293} & 0.246 & \multicolumn{1}{c|}{0.337} & 0.198 & \multicolumn{1}{c|}{0.300} & 0.209 & \multicolumn{1}{c|}{0.301} & 0.217 & \multicolumn{1}{c|}{0.300} & 0.214 & \multicolumn{1}{c|}{0.329} & 0.231 & 0.338 \\
\multicolumn{1}{c|}{\multirow{-4}{*}{Electricity}} & \multicolumn{1}{c|}{720} & {\color[HTML]{FF0000} \textbf{0.212}} & \multicolumn{1}{c|}{{\color[HTML]{FF0000} \textbf{0.293}}} & 0.225 & \multicolumn{1}{c|}{0.317} & {\color[HTML]{00B0F0} {\ul 0.219}} & \multicolumn{1}{c|}{0.315} & 0.225 & \multicolumn{1}{c|}{{\color[HTML]{00B0F0} {\ul 0.310}}} & 0.231 & \multicolumn{1}{c|}{0.332} & 0.231 & \multicolumn{1}{c|}{0.335} & 0.246 & \multicolumn{1}{c|}{0.324} & 0.280 & \multicolumn{1}{c|}{0.363} & 0.220 & \multicolumn{1}{c|}{0.320} & 0.245 & \multicolumn{1}{c|}{0.333} & 0.278 & \multicolumn{1}{c|}{0.356} & 0.246 & \multicolumn{1}{c|}{0.355} & 0.254 & 0.361 \\ \midrule
\multicolumn{1}{c|}{} & \multicolumn{1}{c|}{96} & {\color[HTML]{FF0000} \textbf{0.372}} & \multicolumn{1}{c|}{{\color[HTML]{FF0000} \textbf{0.393}}} & 0.386 & \multicolumn{1}{c|}{0.405} & 0.392 & \multicolumn{1}{c|}{0.407} & 0.385 & \multicolumn{1}{c|}{0.402} & 0.390 & \multicolumn{1}{c|}{0.411} & 0.382 & \multicolumn{1}{c|}{{\color[HTML]{00B0F0} {\ul 0.398}}} & 0.414 & \multicolumn{1}{c|}{0.419} & 0.423 & \multicolumn{1}{c|}{0.448} & 0.384 & \multicolumn{1}{c|}{0.402} & 0.386 & \multicolumn{1}{c|}{0.400} & 0.438 & \multicolumn{1}{c|}{0.433} & {\color[HTML]{00B0F0} {\ul 0.376}} & \multicolumn{1}{c|}{0.419} & 0.449 & 0.459 \\
\multicolumn{1}{c|}{} & \multicolumn{1}{c|}{192} & 0.433 & \multicolumn{1}{c|}{{\color[HTML]{FF0000} \textbf{0.417}}} & 0.441 & \multicolumn{1}{c|}{0.436} & 0.440 & \multicolumn{1}{c|}{0.426} & 0.443 & \multicolumn{1}{c|}{0.430} & 0.442 & \multicolumn{1}{c|}{0.442} & {\color[HTML]{00B0F0} {\ul 0.427}} & \multicolumn{1}{c|}{{\color[HTML]{00B0F0} {\ul 0.425}}} & 0.460 & \multicolumn{1}{c|}{0.445} & 0.471 & \multicolumn{1}{c|}{0.474} & 0.436 & \multicolumn{1}{c|}{0.429} & 0.437 & \multicolumn{1}{c|}{0.432} & 0.493 & \multicolumn{1}{c|}{0.466} & {\color[HTML]{FF0000} \textbf{0.420}} & \multicolumn{1}{c|}{0.448} & 0.500 & 0.482 \\
\multicolumn{1}{c|}{} & \multicolumn{1}{c|}{336} & {\color[HTML]{FF0000} \textbf{0.422}} & \multicolumn{1}{c|}{{\color[HTML]{FF0000} \textbf{0.433}}} & 0.487 & \multicolumn{1}{c|}{0.458} & 0.480 & \multicolumn{1}{c|}{0.453} & 0.512 & \multicolumn{1}{c|}{0.470} & 0.480 & \multicolumn{1}{c|}{0.468} & 0.465 & \multicolumn{1}{c|}{{\color[HTML]{00B0F0} {\ul 0.445}}} & 0.501 & \multicolumn{1}{c|}{0.466} & 0.570 & \multicolumn{1}{c|}{0.546} & 0.491 & \multicolumn{1}{c|}{0.469} & 0.481 & \multicolumn{1}{c|}{0.459} & 0.547 & \multicolumn{1}{c|}{0.495} & {\color[HTML]{00B0F0} {\ul 0.459}} & \multicolumn{1}{c|}{0.465} & 0.521 & 0.496 \\
\multicolumn{1}{c|}{\multirow{-4}{*}{ETTh1}} & \multicolumn{1}{c|}{720} & {\color[HTML]{FF0000} \textbf{0.445}} & \multicolumn{1}{c|}{{\color[HTML]{FF0000} \textbf{0.459}}} & 0.503 & \multicolumn{1}{c|}{0.491} & 0.508 & \multicolumn{1}{c|}{0.493} & 0.498 & \multicolumn{1}{c|}{0.476} & 0.494 & \multicolumn{1}{c|}{0.488} & {\color[HTML]{00B0F0} {\ul 0.472}} & \multicolumn{1}{c|}{{\color[HTML]{00B0F0} {\ul 0.468}}} & 0.500 & \multicolumn{1}{c|}{0.488} & 0.653 & \multicolumn{1}{c|}{0.621} & 0.521 & \multicolumn{1}{c|}{0.500} & 0.519 & \multicolumn{1}{c|}{0.516} & 0.586 & \multicolumn{1}{c|}{0.538} & 0.506 & \multicolumn{1}{c|}{0.507} & 0.514 & 0.512 \\ \midrule
\multicolumn{1}{c|}{} & \multicolumn{1}{c|}{96} & {\color[HTML]{FF0000} \textbf{0.283}} & \multicolumn{1}{c|}{{\color[HTML]{FF0000} \textbf{0.332}}} & 0.297 & \multicolumn{1}{c|}{0.349} & 0.300 & \multicolumn{1}{c|}{0.351} & {\color[HTML]{00B0F0} {\ul 0.296}} & \multicolumn{1}{c|}{{\color[HTML]{00B0F0} {\ul 0.347}}} & 0.328 & \multicolumn{1}{c|}{0.371} & 0.309 & \multicolumn{1}{c|}{0.359} & 0.302 & \multicolumn{1}{c|}{0.348} & 0.745 & \multicolumn{1}{c|}{0.584} & 0.340 & \multicolumn{1}{c|}{0.374} & 0.333 & \multicolumn{1}{c|}{0.387} & 0.322 & \multicolumn{1}{c|}{0.364} & 0.358 & \multicolumn{1}{c|}{0.397} & 0.346 & 0.388 \\
\multicolumn{1}{c|}{} & \multicolumn{1}{c|}{192} & {\color[HTML]{FF0000} \textbf{0.358}} & \multicolumn{1}{c|}{{\color[HTML]{FF0000} \textbf{0.374}}} & 0.380 & \multicolumn{1}{c|}{0.400} & 0.384 & \multicolumn{1}{c|}{0.400} & {\color[HTML]{00B0F0} {\ul 0.376}} & \multicolumn{1}{c|}{{\color[HTML]{00B0F0} {\ul 0.394}}} & 0.402 & \multicolumn{1}{c|}{0.414} & 0.390 & \multicolumn{1}{c|}{0.406} & 0.388 & \multicolumn{1}{c|}{0.400} & 0.877 & \multicolumn{1}{c|}{0.656} & 0.402 & \multicolumn{1}{c|}{0.414} & 0.477 & \multicolumn{1}{c|}{0.476} & 0.404 & \multicolumn{1}{c|}{0.414} & 0.429 & \multicolumn{1}{c|}{0.439} & 0.456 & 0.452 \\
\multicolumn{1}{c|}{} & \multicolumn{1}{c|}{336} & {\color[HTML]{00B0F0} {\ul 0.428}} & \multicolumn{1}{c|}{0.437} & {\color[HTML]{00B0F0} {\ul 0.428}} & \multicolumn{1}{c|}{{\color[HTML]{FF0000} \textbf{0.432}}} & 0.443 & \multicolumn{1}{c|}{0.438} & 0.434 & \multicolumn{1}{c|}{0.443} & 0.435 & \multicolumn{1}{c|}{0.443} & {\color[HTML]{FF0000} \textbf{0.426}} & \multicolumn{1}{c|}{0.444} & {\color[HTML]{FF0000} \textbf{0.426}} & \multicolumn{1}{c|}{{\color[HTML]{00B0F0} {\ul 0.433}}} & 1.043 & \multicolumn{1}{c|}{0.731} & 0.452 & \multicolumn{1}{c|}{0.452} & 0.594 & \multicolumn{1}{c|}{0.541} & 0.435 & \multicolumn{1}{c|}{0.445} & 0.496 & \multicolumn{1}{c|}{0.487} & 0.482 & 0.486 \\
\multicolumn{1}{c|}{\multirow{-4}{*}{ETTh2}} & \multicolumn{1}{c|}{720} & {\color[HTML]{00B0F0} {\ul 0.413}} & \multicolumn{1}{c|}{{\color[HTML]{FF0000} \textbf{0.432}}} & 0.427 & \multicolumn{1}{c|}{0.445} & {\color[HTML]{FF0000} \textbf{0.412}} & \multicolumn{1}{c|}{{\color[HTML]{00B0F0} {\ul 0.441}}} & 0.464 & \multicolumn{1}{c|}{0.464} & 0.417 & \multicolumn{1}{c|}{{\color[HTML]{00B0F0} {\ul 0.441}}} & 0.445 & \multicolumn{1}{c|}{0.464} & 0.431 & \multicolumn{1}{c|}{0.446} & 1.104 & \multicolumn{1}{c|}{0.763} & 0.462 & \multicolumn{1}{c|}{0.468} & 0.831 & \multicolumn{1}{c|}{0.657} & 0.447 & \multicolumn{1}{c|}{0.458} & 0.463 & \multicolumn{1}{c|}{0.474} & 0.515 & 0.511 \\ \midrule
\multicolumn{1}{c|}{} & \multicolumn{1}{c|}{96} & {\color[HTML]{FF0000} \textbf{0.301}} & \multicolumn{1}{c|}{{\color[HTML]{FF0000} \textbf{0.354}}} & 0.334 & \multicolumn{1}{c|}{0.368} & 0.348 & \multicolumn{1}{c|}{0.369} & {\color[HTML]{00B0F0} {\ul 0.318}} & \multicolumn{1}{c|}{{\color[HTML]{00B0F0} {\ul 0.356}}} & 0.319 & \multicolumn{1}{c|}{0.366} & 0.335 & \multicolumn{1}{c|}{0.373} & 0.329 & \multicolumn{1}{c|}{0.367} & 0.404 & \multicolumn{1}{c|}{0.426} & 0.338 & \multicolumn{1}{c|}{0.375} & 0.345 & \multicolumn{1}{c|}{0.372} & 0.353 & \multicolumn{1}{c|}{0.370} & 0.379 & \multicolumn{1}{c|}{0.419} & 0.505 & 0.475 \\
\multicolumn{1}{c|}{} & \multicolumn{1}{c|}{192} & {\color[HTML]{FF0000} \textbf{0.345}} & \multicolumn{1}{c|}{{\color[HTML]{FF0000} \textbf{0.375}}} & 0.377 & \multicolumn{1}{c|}{0.391} & 0.392 & \multicolumn{1}{c|}{0.391} & {\color[HTML]{00B0F0} {\ul 0.366}} & \multicolumn{1}{c|}{{\color[HTML]{00B0F0} {\ul 0.385}}} & 0.376 & \multicolumn{1}{c|}{0.397} & 0.372 & \multicolumn{1}{c|}{0.390} & 0.367 & \multicolumn{1}{c|}{{\color[HTML]{00B0F0} {\ul 0.385}}} & 0.450 & \multicolumn{1}{c|}{0.451} & 0.374 & \multicolumn{1}{c|}{0.387} & 0.380 & \multicolumn{1}{c|}{0.389} & 0.389 & \multicolumn{1}{c|}{0.387} & 0.426 & \multicolumn{1}{c|}{0.441} & 0.553 & 0.496 \\
\multicolumn{1}{c|}{} & \multicolumn{1}{c|}{336} & {\color[HTML]{FF0000} \textbf{0.375}} & \multicolumn{1}{c|}{{\color[HTML]{FF0000} \textbf{0.397}}} & 0.426 & \multicolumn{1}{c|}{0.420} & 0.426 & \multicolumn{1}{c|}{0.410} & {\color[HTML]{00B0F0} {\ul 0.396}} & \multicolumn{1}{c|}{{\color[HTML]{00B0F0} {\ul 0.404}}} & 0.417 & \multicolumn{1}{c|}{0.422} & 0.403 & \multicolumn{1}{c|}{0.411} & 0.399 & \multicolumn{1}{c|}{0.410} & 0.532 & \multicolumn{1}{c|}{0.515} & 0.410 & \multicolumn{1}{c|}{0.411} & 0.413 & \multicolumn{1}{c|}{0.413} & 0.421 & \multicolumn{1}{c|}{0.408} & 0.445 & \multicolumn{1}{c|}{0.459} & 0.621 & 0.537 \\
\multicolumn{1}{c|}{\multirow{-4}{*}{ETTm1}} & \multicolumn{1}{c|}{720} & {\color[HTML]{FF0000} \textbf{0.437}} & \multicolumn{1}{c|}{{\color[HTML]{FF0000} \textbf{0.435}}} & 0.491 & \multicolumn{1}{c|}{0.459} & 0.483 & \multicolumn{1}{c|}{0.448} & {\color[HTML]{00B0F0} {\ul 0.454}} & \multicolumn{1}{c|}{0.441} & 0.481 & \multicolumn{1}{c|}{0.458} & 0.461 & \multicolumn{1}{c|}{0.442} & {\color[HTML]{00B0F0} {\ul 0.454}} & \multicolumn{1}{c|}{{\color[HTML]{00B0F0} {\ul 0.439}}} & 0.666 & \multicolumn{1}{c|}{0.589} & 0.478 & \multicolumn{1}{c|}{0.450} & 0.474 & \multicolumn{1}{c|}{0.453} & 0.481 & \multicolumn{1}{c|}{0.441} & 0.543 & \multicolumn{1}{c|}{0.490} & 0.671 & 0.561 \\ \midrule
\multicolumn{1}{c|}{} & \multicolumn{1}{c|}{96} & {\color[HTML]{FF0000} \textbf{0.165}} & \multicolumn{1}{c|}{{\color[HTML]{FF0000} \textbf{0.257}}} & 0.180 & \multicolumn{1}{c|}{0.264} & 0.183 & \multicolumn{1}{c|}{0.267} & {\color[HTML]{00B0F0} {\ul 0.175}} & \multicolumn{1}{c|}{{\color[HTML]{00B0F0} {\ul 0.258}}} & 0.177 & \multicolumn{1}{c|}{0.262} & 0.176 & \multicolumn{1}{c|}{0.266} & {\color[HTML]{00B0F0} {\ul 0.175}} & \multicolumn{1}{c|}{0.259} & 0.287 & \multicolumn{1}{c|}{0.366} & 0.187 & \multicolumn{1}{c|}{0.267} & 0.193 & \multicolumn{1}{c|}{0.292} & 0.183 & \multicolumn{1}{c|}{0.266} & 0.203 & \multicolumn{1}{c|}{0.287} & 0.255 & 0.339 \\
\multicolumn{1}{c|}{} & \multicolumn{1}{c|}{192} & {\color[HTML]{FF0000} \textbf{0.230}} & \multicolumn{1}{c|}{0.307} & 0.250 & \multicolumn{1}{c|}{0.309} & 0.257 & \multicolumn{1}{c|}{0.313} & 0.241 & \multicolumn{1}{c|}{{\color[HTML]{00B0F0} {\ul 0.304}}} & 0.247 & \multicolumn{1}{c|}{0.307} & {\color[HTML]{00B0F0} {\ul 0.240}} & \multicolumn{1}{c|}{0.307} & 0.241 & \multicolumn{1}{c|}{{\color[HTML]{FF0000} \textbf{0.302}}} & 0.414 & \multicolumn{1}{c|}{0.492} & 0.249 & \multicolumn{1}{c|}{0.309} & 0.284 & \multicolumn{1}{c|}{0.362} & 0.248 & \multicolumn{1}{c|}{0.305} & 0.269 & \multicolumn{1}{c|}{0.328} & 0.281 & 0.340 \\
\multicolumn{1}{c|}{} & \multicolumn{1}{c|}{336} & {\color[HTML]{FF0000} \textbf{0.282}} & \multicolumn{1}{c|}{{\color[HTML]{FF0000} \textbf{0.328}}} & 0.311 & \multicolumn{1}{c|}{0.348} & 0.335 & \multicolumn{1}{c|}{0.361} & {\color[HTML]{00B0F0} {\ul 0.303}} & \multicolumn{1}{c|}{{\color[HTML]{00B0F0} {\ul 0.343}}} & 0.312 & \multicolumn{1}{c|}{0.346} & 0.304 & \multicolumn{1}{c|}{0.345} & 0.305 & \multicolumn{1}{c|}{{\color[HTML]{00B0F0} {\ul 0.343}}} & 0.597 & \multicolumn{1}{c|}{0.542} & 0.321 & \multicolumn{1}{c|}{0.351} & 0.369 & \multicolumn{1}{c|}{0.427} & 0.309 & \multicolumn{1}{c|}{{\color[HTML]{00B0F0} {\ul 0.343}}} & 0.325 & \multicolumn{1}{c|}{0.366} & 0.339 & 0.372 \\
\multicolumn{1}{c|}{\multirow{-4}{*}{ETTm2}} & \multicolumn{1}{c|}{720} & {\color[HTML]{FF0000} \textbf{0.375}} & \multicolumn{1}{c|}{{\color[HTML]{00B0F0} {\ul 0.396}}} & 0.412 & \multicolumn{1}{c|}{0.407} & 0.410 & \multicolumn{1}{c|}{0.402} & {\color[HTML]{00B0F0} {\ul 0.391}} & \multicolumn{1}{c|}{{\color[HTML]{FF0000} \textbf{0.394}}} & 0.414 & \multicolumn{1}{c|}{0.403} & 0.406 & \multicolumn{1}{c|}{0.400} & 0.402 & \multicolumn{1}{c|}{0.400} & 1.730 & \multicolumn{1}{c|}{1.042} & 0.408 & \multicolumn{1}{c|}{0.403} & 0.554 & \multicolumn{1}{c|}{0.522} & 0.410 & \multicolumn{1}{c|}{0.400} & 0.421 & \multicolumn{1}{c|}{0.415} & 0.433 & 0.432 \\ \midrule
\multicolumn{1}{c|}{} & \multicolumn{1}{c|}{96} & {\color[HTML]{FF0000} \textbf{0.384}} & \multicolumn{1}{c|}{{\color[HTML]{FF0000} \textbf{0.248}}} & {\color[HTML]{00B0F0} {\ul 0.395}} & \multicolumn{1}{c|}{{\color[HTML]{00B0F0} {\ul 0.268}}} & 0.413 & \multicolumn{1}{c|}{0.272} & 0.473 & \multicolumn{1}{c|}{0.288} & 0.605 & \multicolumn{1}{c|}{0.344} & 0.570 & \multicolumn{1}{c|}{0.310} & 0.462 & \multicolumn{1}{c|}{0.295} & 0.522 & \multicolumn{1}{c|}{0.290} & 0.593 & \multicolumn{1}{c|}{0.321} & 0.650 & \multicolumn{1}{c|}{0.396} & 0.647 & \multicolumn{1}{c|}{0.384} & 0.587 & \multicolumn{1}{c|}{0.366} & 0.613 & 0.388 \\
\multicolumn{1}{c|}{} & \multicolumn{1}{c|}{192} & {\color[HTML]{FF0000} \textbf{0.401}} & \multicolumn{1}{c|}{{\color[HTML]{FF0000} \textbf{0.258}}} & {\color[HTML]{00B0F0} {\ul 0.417}} & \multicolumn{1}{c|}{0.276} & 0.422 & \multicolumn{1}{c|}{{\color[HTML]{00B0F0} {\ul 0.274}}} & 0.473 & \multicolumn{1}{c|}{0.296} & 0.613 & \multicolumn{1}{c|}{0.359} & 0.577 & \multicolumn{1}{c|}{0.321} & 0.466 & \multicolumn{1}{c|}{0.296} & 0.530 & \multicolumn{1}{c|}{0.293} & 0.617 & \multicolumn{1}{c|}{0.336} & 0.598 & \multicolumn{1}{c|}{0.370} & 0.600 & \multicolumn{1}{c|}{0.361} & 0.604 & \multicolumn{1}{c|}{0.373} & 0.616 & 0.382 \\
\multicolumn{1}{c|}{} & \multicolumn{1}{c|}{336} & {\color[HTML]{FF0000} \textbf{0.423}} & \multicolumn{1}{c|}{{\color[HTML]{FF0000} \textbf{0.261}}} & {\color[HTML]{00B0F0} {\ul 0.433}} & \multicolumn{1}{c|}{{\color[HTML]{00B0F0} {\ul 0.283}}} & 0.438 & \multicolumn{1}{c|}{0.292} & 0.508 & \multicolumn{1}{c|}{0.312} & 0.642 & \multicolumn{1}{c|}{0.376} & 0.588 & \multicolumn{1}{c|}{0.324} & 0.482 & \multicolumn{1}{c|}{0.304} & 0.558 & \multicolumn{1}{c|}{0.305} & 0.629 & \multicolumn{1}{c|}{0.336} & 0.605 & \multicolumn{1}{c|}{0.373} & 0.610 & \multicolumn{1}{c|}{0.367} & 0.621 & \multicolumn{1}{c|}{0.383} & 0.622 & 0.337 \\
\multicolumn{1}{c|}{\multirow{-4}{*}{Traffic}} & \multicolumn{1}{c|}{720} & {\color[HTML]{FF0000} \textbf{0.453}} & \multicolumn{1}{c|}{{\color[HTML]{FF0000} \textbf{0.282}}} & 0.467 & \multicolumn{1}{c|}{0.302} & {\color[HTML]{00B0F0} {\ul 0.457}} & \multicolumn{1}{c|}{{\color[HTML]{00B0F0} {\ul 0.292}}} & 0.512 & \multicolumn{1}{c|}{0.318} & 0.702 & \multicolumn{1}{c|}{0.401} & 0.597 & \multicolumn{1}{c|}{0.337} & 0.514 & \multicolumn{1}{c|}{0.322} & 0.589 & \multicolumn{1}{c|}{0.328} & 0.640 & \multicolumn{1}{c|}{0.350} & 0.645 & \multicolumn{1}{c|}{0.394} & 0.691 & \multicolumn{1}{c|}{0.425} & 0.626 & \multicolumn{1}{c|}{0.382} & 0.660 & 0.408 \\ \bottomrule
\label{tab:addlabel1}
\end{tabular}}
\end{table}

\textbf{Long-Range Time Series Forecasting Under Univariate Settings.} Table \ref{tab:addlabel2} and Table \ref{tab:addlabel3} summarize the average results and full results of long-range time series forecasting under univariate settings, where the results of baselines without * are cited from DLinear \cite{Dlinear}. Following existing works \cite{FEDformer, MSHyper, PatchTST, Autoformer}, we set the univariate forecasting on ETT as only predicting a target variate "oil temperature" given inputs from all variables. We can see from Table \ref{tab:addlabel2} that Ada-MSHyper achieves the SOTA results on all datasets. Specifically, Ada-MSHyper gives an average error reduction of 7.57\% and 4.65\% compared to the best baseline in terms of MSE and MAE, respectively. 

\begin{table}[htbp]
\caption{Long-range time series forecasting results under univariate settings.}
\resizebox{\linewidth}{!}{
\begin{tabular}{@{}ccccccccccccc@{}}
\toprule
\multicolumn{2}{c}{Models} & \begin{tabular}[c]{@{}c@{}}Ada-MSHyper\\      (Ours)\end{tabular} & \begin{tabular}[c]{@{}c@{}}iTransfomer*\\      (2024)\end{tabular} & \begin{tabular}[c]{@{}c@{}}MSHyper*\\      (2024)\end{tabular} & \begin{tabular}[c]{@{}c@{}}TimeMixer*\\      (2024)\end{tabular} & \begin{tabular}[c]{@{}c@{}}PatchTST*\\      (2023)\end{tabular} & \begin{tabular}[c]{@{}c@{}}DLinear\\      (2023)\end{tabular} & \begin{tabular}[c]{@{}c@{}}Crossformer*\\      (2023)\end{tabular} & \begin{tabular}[c]{@{}c@{}}Pyraformer\\      (2022)\end{tabular} & \begin{tabular}[c]{@{}c@{}}FEDformer\\      (2022)\end{tabular} & \begin{tabular}[c]{@{}c@{}}Autoformer\\      (2021)\end{tabular} & \begin{tabular}[c]{@{}c@{}}Informer\\      (2021)\end{tabular} \\ \midrule
\multicolumn{2}{c|}{Metric} & \multicolumn{1}{c|}{MSE MAE} & \multicolumn{1}{c|}{MSE MAE} & \multicolumn{1}{c|}{MSE MAE} & \multicolumn{1}{c|}{MSE MAE} & \multicolumn{1}{c|}{MSE MAE} & \multicolumn{1}{c|}{MSE MAE} & \multicolumn{1}{c|}{MSE MAE} & \multicolumn{1}{c|}{MSE MAE} & \multicolumn{1}{c|}{MSE MAE} & \multicolumn{1}{c|}{MSE MAE} & MSE MAE \\ \midrule
\multicolumn{2}{c|}{ETTh1} & \multicolumn{1}{c|}{{\color[HTML]{FF0000} \textbf{0.071 0.203}}} & \multicolumn{1}{c|}{0.076 0.213} & \multicolumn{1}{c|}{0.080 0.218} & \multicolumn{1}{c|}{{\color[HTML]{00CCFF} {\ul 0.074 } \color[HTML]{00CCFF} {\ul 0.210}}} & \multicolumn{1}{c|}{0.082 0.221} & \multicolumn{1}{c|}{0.104 0.247} & \multicolumn{1}{c|}{0.099 0.246} & \multicolumn{1}{c|}{0.170 0.335} & \multicolumn{1}{c|}{0.111 0.257} & \multicolumn{1}{c|}{0.105 0.252} & 0.199 0.377 \\ \midrule
\multicolumn{2}{c|}{ETTh2} & \multicolumn{1}{c|}{{\color[HTML]{FF0000} \textbf{0.171 0.329}}} & \multicolumn{1}{c|}{0.199 0.352} & \multicolumn{1}{c|}{{\color[HTML]{00CCFF} {\ul 0.187} \color[HTML]{00CCFF} {\ul 0.338}}} & \multicolumn{1}{c|}{0.198 0.350} & \multicolumn{1}{c|}{0.198 0.348} & \multicolumn{1}{c|}{0.198 0.350} & \multicolumn{1}{c|}{0.201 0.349} & \multicolumn{1}{c|}{0.215 0.373} & \multicolumn{1}{c|}{0.206 0.350} & \multicolumn{1}{c|}{0.218 0.364} & 0.243 0.400 \\ \midrule
\multicolumn{2}{c|}{ETTm1} & \multicolumn{1}{c|}{{\color[HTML]{FF0000} \textbf{0.047 0.159}}} &\multicolumn{1}{c|}{ 0.053 0.174 }&\multicolumn{1}{c|}{ 0.053 0.172 }&\multicolumn{1}{c|}{ 0.053 0.173 }&\multicolumn{1}{c|}{ {\color[HTML]{00CCFF} {\ul 0.052}} 0.171 }&\multicolumn{1}{c|}{ 0.054 \color[HTML]{00CCFF} {\ul 0.168} }&\multicolumn{1}{c|}{ 0.065 0.196 }&\multicolumn{1}{c|}{ 0.255 0.392 }&\multicolumn{1}{c|}{ 0.069 0.202 }&\multicolumn{1}{c|}{ 0.081 0.221 }& 0.281 0.441 \\ \midrule
\multicolumn{2}{c|}{ETTm2} &\multicolumn{1}{c|}{ {\color[HTML]{FF0000} \textbf{0.103 0.230}} }&\multicolumn{1}{c|}{ 0.128 0.268 }&\multicolumn{1}{c|}{ 0.120 0.258 }&\multicolumn{1}{c|}{ 0.121 0.258 }&\multicolumn{1}{c|}{ 0.121 0.258 }&\multicolumn{1}{c|}{ {\color[HTML]{00CCFF} {\ul 0.112} \color[HTML]{00CCFF} {\ul 0.248}} }&\multicolumn{1}{c|}{ 0.119 0.256 }&\multicolumn{1}{c|}{ 0.133 0.273 }&\multicolumn{1}{c|}{ 0.119 0.262 }&\multicolumn{1}{c|}{ 0.130 0.271 }& 0.175 0.320 \\ \bottomrule
\label{tab:addlabel2}
\end{tabular}}
\end{table}

\begin{table}[htbp]
\caption{Full results of long-range time series forecasting under univariate settings.}
\begin{tiny}
\resizebox{\linewidth}{!}{
\begin{tabular}{@{}ccccccccccccc@{}}
\toprule
\multicolumn{2}{c}{Models} & \begin{tabular}[c]{@{}c@{}}Ada-MSHyper\\      (Ours)\end{tabular} & \begin{tabular}[c]{@{}c@{}}iTransfomer*\\      (2024)\end{tabular} & \begin{tabular}[c]{@{}c@{}}MSHyper*\\      (2024)\end{tabular} & \begin{tabular}[c]{@{}c@{}}TimeMixer*\\      (2024)\end{tabular} & \begin{tabular}[c]{@{}c@{}}PatchTST*\\      (2023)\end{tabular} & \begin{tabular}[c]{@{}c@{}}DLinear\\      (2023)\end{tabular} & \begin{tabular}[c]{@{}c@{}}Crossformer\\      (2023)\end{tabular} & \begin{tabular}[c]{@{}c@{}}Pyraformer\\      (2022)\end{tabular} & \begin{tabular}[c]{@{}c@{}}FEDformer\\      (2022)\end{tabular} & \begin{tabular}[c]{@{}c@{}}Autoformer\\      (2021)\end{tabular} & \begin{tabular}[c]{@{}c@{}}Informer\\      (2021)\end{tabular} \\ \midrule
\multicolumn{2}{c|}{Metric} & \multicolumn{1}{c|}{MSE MAE} & \multicolumn{1}{c|}{MSE MAE} &\multicolumn{1}{c|}{ MSE MAE} & \multicolumn{1}{c|}{MSE MAE} & \multicolumn{1}{c|}{MSE MAE} &\multicolumn{1}{c|}{ MSE MAE} & \multicolumn{1}{c|}{MSE MAE} & \multicolumn{1}{c|}{MSE MAE} & \multicolumn{1}{c|}{MSE MAE} & \multicolumn{1}{c|}{MSE MAE} & \multicolumn{1}{c}{MSE MAE} \\ \midrule
\multicolumn{1}{c|}{} & \multicolumn{1}{c|}{96} & \multicolumn{1}{c|}{{\color[HTML]{00CCFF} {\ul \textbf{0.057}} \color[HTML]{FF0000} \textbf{0.173}}} & \multicolumn{1}{c|}{0.059 0.185} & \multicolumn{1}{c|}{{\color[HTML]{FF0000} \textbf{0.056}} 0.181} & \multicolumn{1}{c|}{{\color[HTML]{00CCFF} 0.057} 0.181}& \multicolumn{1}{c|}{{\color[HTML]{FF0000} \textbf{0.056}} 0.181} & \multicolumn{1}{c|}{{\color[HTML]{FF0000} \textbf{0.056}} \color[HTML]{00CCFF} {\ul 0.180}} & \multicolumn{1}{c|}{0.076 0.216} & \multicolumn{1}{c|}{0.099 0.277} & \multicolumn{1}{c|}{0.079 0.215} & \multicolumn{1}{c|}{0.071 0.206} & \multicolumn{1}{c}{0.193 0.377} \\
\multicolumn{1}{c|}{} & \multicolumn{1}{c|}{192} & \multicolumn{1}{c|}{{\color[HTML]{00CCFF} {\ul \textbf{0.072}} \color[HTML]{FF0000} \textbf{0.198}}} & \multicolumn{1}{c|}{0.073 0.208} & \multicolumn{1}{c|}{0.076 0.211} & {\color[HTML]{00CCFF} {\ul 0.072}}  {\color[HTML]{00CCFF} {\ul 0.204}}& \multicolumn{1}{c|}{0.076 0.210} & \multicolumn{1}{c|}{{\color[HTML]{FF0000} \textbf{0.071} \color[HTML]{00CCFF} {\ul 0.204}}} & \multicolumn{1}{c|}{0.085 0.225} & \multicolumn{1}{c|}{0.174 0.346} & \multicolumn{1}{c|}{0.104 0.245} & \multicolumn{1}{c|}{0.114 0.262 }& \multicolumn{1}{c}{0.217 0.395} \\
\multicolumn{1}{c|}{} & \multicolumn{1}{c|}{336} & \multicolumn{1}{c|}{{\color[HTML]{FF0000} \textbf{0.070 0.213}}} & \multicolumn{1}{c|}{{\color[HTML]{00CCFF} {\ul 0.084} \color[HTML]{00CCFF} {\ul0.223}}} & \multicolumn{1}{c|}{0.090 0.236} & \multicolumn{1}{c|}{0.085 0.227} & \multicolumn{1}{c|}{0.094 0.242} & \multicolumn{1}{c|}{0.098 0.244} & \multicolumn{1}{c|}{0.106 0.257} & \multicolumn{1}{c|}{0.198 0.370} &\multicolumn{1}{c|} {0.119 0.270} & \multicolumn{1}{c|}{0.107 0.258} & \multicolumn{1}{c}{0.202 0.381} \\
\multicolumn{1}{c|}{\multirow{-4}{*}{ETTh1}} & \multicolumn{1}{c|}{720} & \multicolumn{1}{c|}{{\color[HTML]{00CCFF} {\color[HTML]{00CCFF} {\ul 0.085}} \color[HTML]{00CCFF} {\ul 0.228}}} & \multicolumn{1}{c|}{0.089 0.236} & \multicolumn{1}{c|}{0.096 0.245} & \multicolumn{1}{c|}{{\color[HTML]{FF0000} \textbf{0.083 0.227}} }& \multicolumn{1}{c|}{0.101 0.250} & \multicolumn{1}{c|}{0.189 0.359} & \multicolumn{1}{c|}{0.128 0.287} & \multicolumn{1}{c|}{0.209 0.348} & \multicolumn{1}{c|}{0.142 0.299} & \multicolumn{1}{c|}{0.126 0.283} &\multicolumn{1}{c} {0.183 0.355} \\ \midrule
\multicolumn{1}{c|}{} & \multicolumn{1}{c|}{96} & \multicolumn{1}{c|}{{\color[HTML]{FF0000} \textbf{0.116 0.262}}} & \multicolumn{1}{c|}{0.136 0.287} & \multicolumn{1}{c|}{{\color[HTML]{00CCFF} {\ul 0.117} \color[HTML]{00CCFF} {\ul0.266}}} &\multicolumn{1}{c|}{ 0.133 0.283} & \multicolumn{1}{c|}{0.130 0.276} & \multicolumn{1}{c|}{0.131 0.279} & \multicolumn{1}{c|}{0.125 0.273} & \multicolumn{1}{c|}{0.152 0.303} & \multicolumn{1}{c|}{0.128 0.271} & \multicolumn{1}{c|}{0.153 0.306} & \multicolumn{1}{c}{0.213 0.373} \\
\multicolumn{1}{c|}{} & \multicolumn{1}{c|}{192} & \multicolumn{1}{c|}{{\color[HTML]{FF0000} \textbf{0.168 0.323}}} & \multicolumn{1}{c|}{0.187 0.342} & \multicolumn{1}{c|}{{\color[HTML]{00CCFF} {\ul 0.172} \color[HTML]{00CCFF} {\ul 0.325}}} & \multicolumn{1}{c|}{0.190 0.341} &\multicolumn{1}{c|} {0.181 0.331} & \multicolumn{1}{c|}{0.176 0.329} & \multicolumn{1}{c|}{0.187 0.334} & \multicolumn{1}{c|}{0.197 0.370} &  \multicolumn{1}{c|}{0.185 0.330} & \multicolumn{1}{c|}{0.204 0.351} & \multicolumn{1}{c}{0.227 0.387} \\
\multicolumn{1}{c|}{} & \multicolumn{1}{c|}{336} & \multicolumn{1}{c|}{{\color[HTML]{FF0000} \textbf{0.177 0.350}}} &\multicolumn{1}{c|} {0.219 0.374} & \multicolumn{1}{c|}{0.211 \color[HTML]{00CCFF} {\ul 0.362}} & \multicolumn{1}{c|}{0.226 0.379} & \multicolumn{1}{c|}{0.226 0.379} & \multicolumn{1}{c|}{{\color[HTML]{00CCFF} {\ul 0.209}} 0.367} & \multicolumn{1}{c|}{0.227 0.377} &\multicolumn{1}{c|}{ 0.238 0.385} & \multicolumn{1}{c|}{0.231 0.378} & \multicolumn{1}{c|}{0.246 0.389} & \multicolumn{1}{c}{0.242 0.401} \\
\multicolumn{1}{c|}{\multirow{-4}{*}{ETTh2}} & \multicolumn{1}{c|}{720} & \multicolumn{1}{c|}{{\color[HTML]{FF0000} \textbf{0.221 0.380}}} & \multicolumn{1}{c|}{0.253 0.403} & \multicolumn{1}{c|}{0.248 0.398} & \multicolumn{1}{c|}{{\color[HTML]{00CCFF} {\ul 0.241}} {\color[HTML]{00CCFF} {\ul 0.396}}} & \multicolumn{1}{c|}{0.253 0.406} & \multicolumn{1}{c|}{0.276 0.426} & \multicolumn{1}{c|}{0.266 0.410} & \multicolumn{1}{c|}{0.274 0.435} & \multicolumn{1}{c|}{0.278 0.420} & \multicolumn{1}{c|}{0.268 0.409} & \multicolumn{1}{c}{0.291 0.439} \\ \midrule
\multicolumn{1}{c|}{} & \multicolumn{1}{c|}{96} & \multicolumn{1}{c|}{{\color[HTML]{FF0000} \textbf{0.027 0.118}}} & \multicolumn{1}{c|}{0.029 0.127} & \multicolumn{1}{c|}{0.029 0.127} & \multicolumn{1}{c|}{0.029 0.128} & \multicolumn{1}{c|}{0.029 0.126} & \multicolumn{1}{c|}{{\color[HTML]{00CCFF} {\ul 0.028}} {\color[HTML]{00CCFF} {\ul 0.123}}} & \multicolumn{1}{c|}{0.035 0.145} & \multicolumn{1}{c|}{0.127 0.281} & \multicolumn{1}{c|}{0.033 0.140} & \multicolumn{1}{c|}{0.056 0.183} & \multicolumn{1}{c}{0.109 0.277} \\
\multicolumn{1}{c|}{} & \multicolumn{1}{c|}{192} & \multicolumn{1}{c|}{{\color[HTML]{FF0000} \textbf{0.038 0.148}}} & \multicolumn{1}{c|}{0.045 0.162} & \multicolumn{1}{c|}{0.044 0.159} & \multicolumn{1}{c|}{0.044 0.160} & \multicolumn{1}{c|}{{\color[HTML]{00CCFF} {\ul 0.043}} 0.158} & \multicolumn{1}{c|}{0.045 {\color[HTML]{00CCFF} {\ul 0.156}}} & \multicolumn{1}{c|}{0.055 0.180} & \multicolumn{1}{c|}{0.205 0.343} & \multicolumn{1}{c|}{0.058 0.186} & \multicolumn{1}{c|}{0.081 0.216} & \multicolumn{1}{c}{0.151 0.310} \\
\multicolumn{1}{c|}{} & \multicolumn{1}{c|}{336} & \multicolumn{1}{c|}{{\color[HTML]{FF0000} \textbf{0.052 0.165}}} & \multicolumn{1}{c|}{0.059 0.189} & \multicolumn{1}{c|}{0.059 0.186} & \multicolumn{1}{c|}{0.058 0.185} & \multicolumn{1}{c|}{{\color[HTML]{00CCFF} {\ul 0.056}} 0.183} & \multicolumn{1}{c|}{0.061 {\color[HTML]{00CCFF} {\ul 0.182}}} & \multicolumn{1}{c|}{0.072 0.209} & \multicolumn{1}{c|}{0.302 0.457} & \multicolumn{1}{c|}{0.084 0.231} & \multicolumn{1}{c|}{0.076 0.218} & \multicolumn{1}{c}{0.427 0.591} \\
\multicolumn{1}{c|}{\multirow{-4}{*}{ETTm1}} & \multicolumn{1}{c|}{720} & \multicolumn{1}{c|}{{\color[HTML]{FF0000} \textbf{0.071 0.206}}} & \multicolumn{1}{c|}{{\color[HTML]{00CCFF} {\ul 0.080}} 0.218} & \multicolumn{1}{c|}{{\color[HTML]{00CCFF} {\ul 0.080}} 0.217} & \multicolumn{1}{c|}{0.081 0.218} & \multicolumn{1}{c|}{{\color[HTML]{00CCFF} {\ul 0.080}} 0.217} & \multicolumn{1}{c|}{{\color[HTML]{00CCFF} {\ul 0.080 }} {\color[HTML]{00CCFF} {\ul 0.210}}} & \multicolumn{1}{c|}{0.097 0.248} & \multicolumn{1}{c|}{0.387 0.485} & \multicolumn{1}{c|}{0.102 0.250} & \multicolumn{1}{c|}{0.110 0.267} & \multicolumn{1}{c}{0.438 0.586} \\ \midrule
\multicolumn{1}{c|}{} & \multicolumn{1}{c|}{96} & \multicolumn{1}{c|}{{\color[HTML]{FF0000} \textbf{0.051 0.163}}} & \multicolumn{1}{c|}{0.071 0.193} & \multicolumn{1}{c|}{0.071 0.194} & \multicolumn{1}{c|}{0.068 0.187} & \multicolumn{1}{c|}{0.071 0.192} & \multicolumn{1}{c|}{0.063 {\color[HTML]{00CCFF} {\ul 0.183}}} & \multicolumn{1}{c|}{{\color[HTML]{00CCFF} {\ul 0.058 0.183}}} & \multicolumn{1}{c|}{0.074 0.208} & \multicolumn{1}{c|}{0.067 0.198} & \multicolumn{1}{c|}{0.065 0.189} & \multicolumn{1}{c}{0.088 0.225} \\
\multicolumn{1}{c|}{} & \multicolumn{1}{c|}{192} & \multicolumn{1}{c|}{{\color[HTML]{FF0000} \textbf{0.089 0.207}}} & \multicolumn{1}{c|}{0.109 0.248} & \multicolumn{1}{c|}{0.102 0.238} & \multicolumn{1}{c|}{0.101 0.236} & \multicolumn{1}{c|}{0.102 0.237} & \multicolumn{1}{c|}{{\color[HTML]{00CCFF} {\ul 0.092 }} {\color[HTML]{00CCFF} {\ul 0.227}}} & \multicolumn{1}{c|}{0.105 0.237} & \multicolumn{1}{c|}{0.116 0.252} & \multicolumn{1}{c|}{0.102 0.245} & \multicolumn{1}{c|}{0.118 0.256} & \multicolumn{1}{c}{0.132 0.283} \\
\multicolumn{1}{c|}{} & \multicolumn{1}{c|}{336} & \multicolumn{1}{c|}{{\color[HTML]{FF0000} \textbf{0.114 0.240}}} & \multicolumn{1}{c|}{0.141 0.289} & \multicolumn{1}{c|}{0.129 0.274} & \multicolumn{1}{c|}{0.133 0.278} & \multicolumn{1}{c|}{0.130 0.274} & \multicolumn{1}{c|}{{\color[HTML]{00CCFF} {\ul 0.119}} {\color[HTML]{00CCFF} {\ul 0.261}}} & \multicolumn{1}{c|}{0.133 0.280} & \multicolumn{1}{c|}{0.143 0.295} & \multicolumn{1}{c|}{0.130 0.279} & \multicolumn{1}{c|}{0.154 0.305} & \multicolumn{1}{c}{0.180 0.336} \\
\multicolumn{1}{c|}{\multirow{-4}{*}{ETTm2}} & \multicolumn{1}{c|}{720} & \multicolumn{1}{c|}{{\color[HTML]{FF0000} \textbf{0.156 0.310}}} & \multicolumn{1}{c|}{0.190 0.343} & \multicolumn{1}{c|}{0.176 0.324} & \multicolumn{1}{c|}{0.183 0.332} & \multicolumn{1}{c|}{0.179 0.328} & \multicolumn{1}{c|}{{\color[HTML]{00CCFF} {\ul 0.175}} {\color[HTML]{00CCFF} {\ul 0.320}}} & \multicolumn{1}{c|}{0.181 0.324} & \multicolumn{1}{c|}{0.197 0.338} & \multicolumn{1}{c|}{0.178 0.325} & \multicolumn{1}{c|}{0.182 0.335} & \multicolumn{1}{c}{0.300 0.435} \\ \bottomrule
\label{tab:addlabel3}
\end{tabular}}
\end{tiny}
\end{table}

\textbf{Short-Range Time Series Forecasting Under Multivariate Settings.}
Table \ref{tab:addlabel4} summarizes the results of short-range time series forecasting under multivariate settings, where the results of baselines without * are cited from iTransformer \cite{Itransformer}. We can see from Table \ref{tab:addlabel4} that Ada-MSHyper achieves the SOTA results on all datasets. Specifically, Ada-MSHyper gives an average error reduction of 10.38\% and 3.82\% compared to the best baseline in terms of MSE and MAE, respectively.

\begin{table}[htbp]
\caption{Full results of short-range time series forecasting under multivariate settings.}
\resizebox{\linewidth}{!}{
\begin{tabular}{@{}ccccccccccccc@{}}
\toprule
\multicolumn{2}{c}{Models} & \begin{tabular}[c]{@{}c@{}}Ada-MSHyper\\      (Ours)\end{tabular} & \begin{tabular}[c]{@{}c@{}}iTransformer*\\      (2024)\end{tabular} & \begin{tabular}[c]{@{}c@{}}MSHyper*\\      (2024)\end{tabular} & \begin{tabular}[c]{@{}c@{}}TimeMixer*\\      (2024)\end{tabular} & \begin{tabular}[c]{@{}c@{}}PatchTST\\      (2023)\end{tabular} & \begin{tabular}[c]{@{}c@{}}TimesNet\\      (2023)\end{tabular} & \begin{tabular}[c]{@{}c@{}}DLinear\\      (2023)\end{tabular} & \begin{tabular}[c]{@{}c@{}}Crossformer\\      (2023)\end{tabular} & \begin{tabular}[c]{@{}c@{}}SCINet\\      (2022)\end{tabular} & \begin{tabular}[c]{@{}c@{}}FEDformer\\      (2022)\end{tabular} & \begin{tabular}[c]{@{}c@{}}Autoformer\\      (2021)\end{tabular} \\ \midrule
\multicolumn{2}{c|}{Metric} & \multicolumn{1}{c|}{MSE MAE} & \multicolumn{1}{c|}{MSE MAE} & \multicolumn{1}{c|}{MSE MAE} & \multicolumn{1}{c|}{MSE MAE} & \multicolumn{1}{c|}{MSE MAE} & \multicolumn{1}{c|}{MSE MAE} & \multicolumn{1}{c|}{MSE MAE} & \multicolumn{1}{c|}{MSE MAE} & \multicolumn{1}{c|}{MSE MAE} & \multicolumn{1}{c|}{MSE MAE} & MSE MAE \\ \midrule
\multicolumn{1}{c|}{} & \multicolumn{1}{c|}{12} & \multicolumn{1}{c|}{{\color[HTML]{FF0000} \textbf{0.060 0.165}}} & \multicolumn{1}{c|}{0.071 0.174} & \multicolumn{1}{c|}{0.106 0.207} & \multicolumn{1}{c|}{0.161 0.323} & \multicolumn{1}{c|}{0.099 0.216} & \multicolumn{1}{c|}{0.085 0.192} & \multicolumn{1}{c|}{0.122 0.243} & \multicolumn{1}{c|}{0.090 0.203} & \multicolumn{1}{c|}{{\color[HTML]{00CCFF} {\ul 0.066}} {\color[HTML]{00CCFF} {\ul 0.172}}} & \multicolumn{1}{c|}{0.126 0.251} & 0.272 0.385 \\
\multicolumn{1}{c|}{} & \multicolumn{1}{c|}{24} & \multicolumn{1}{c|}{{\color[HTML]{FF0000} \textbf{0.075 0.184}}} & \multicolumn{1}{c|}{0.093 0.201} & \multicolumn{1}{c|}{0.126 0.207} & \multicolumn{1}{c|}{0.181 0.352} & \multicolumn{1}{c|}{0.142 0.259} & \multicolumn{1}{c|}{0.118 0.223} & \multicolumn{1}{c|}{0.201 0.317} & \multicolumn{1}{c|}{0.121 0.240} & \multicolumn{1}{c|}{{\color[HTML]{00CCFF} {\ul 0.085}} {\color[HTML]{00CCFF} {\ul 0.198}}} & \multicolumn{1}{c|}{0.149 0.275} & 0.334 0.440 \\
\multicolumn{1}{c|}{\multirow{-3}{*}{PEMS03}} & \multicolumn{1}{c|}{48} & \multicolumn{1}{c|}{{\color[HTML]{FF0000} \textbf{0.120 0.230}}} & \multicolumn{1}{c|}{{\color[HTML]{00CCFF} {\ul 0.125}} {\color[HTML]{00CCFF} {\ul 0.236}}} & \multicolumn{1}{c|}{0.138 0.265} & \multicolumn{1}{c|}{0.222 0.407} & \multicolumn{1}{c|}{0.211 0.319} & \multicolumn{1}{c|}{0.155 0.260} & \multicolumn{1}{c|}{0.333 0.425} & \multicolumn{1}{c|}{0.202 0.317} & \multicolumn{1}{c|}{0.127 0.238} & \multicolumn{1}{c|}{0.227 0.348} & 1.032 0.782 \\ \midrule
\multicolumn{1}{c|}{} & \multicolumn{1}{c|}{12} & \multicolumn{1}{c|}{{\color[HTML]{FF0000} \textbf{0.068 0.173}}} & \multicolumn{1}{c|}{0.078 0.183} & \multicolumn{1}{c|}{0.103 0.197} & \multicolumn{1}{c|}{0.168 0.344} & \multicolumn{1}{c|}{0.105 0.224} & \multicolumn{1}{c|}{0.087 0.195} & \multicolumn{1}{c|}{0.148 0.272} & \multicolumn{1}{c|}{0.098 0.218} & \multicolumn{1}{c|}{{\color[HTML]{00CCFF} {\ul 0.073}} {\color[HTML]{00CCFF} {\ul  0.177}}} & \multicolumn{1}{c|}{0.138 0.262} & 0.424 0.491 \\
\multicolumn{1}{c|}{} & \multicolumn{1}{c|}{24} & \multicolumn{1}{c|}{{\color[HTML]{FF0000} \textbf{0.080 0.189}}} & \multicolumn{1}{c|}{0.095 0.205} & \multicolumn{1}{c|}{0.148 0.245} & \multicolumn{1}{c|}{0.183 0.362} & \multicolumn{1}{c|}{0.153 0.275} & \multicolumn{1}{c|}{0.103 0.215} & \multicolumn{1}{c|}{0.224 0.340} & \multicolumn{1}{c|}{0.131 0.256} & \multicolumn{1}{c|}{{\color[HTML]{00CCFF} {\ul 0.084}} {\color[HTML]{00CCFF} {\ul }} {\color[HTML]{00CCFF} {\ul  0.193}}} & \multicolumn{1}{c|}{0.177 0.293} & 0.459 0.509 \\
\multicolumn{1}{c|}{\multirow{-3}{*}{PEMS04}} & \multicolumn{1}{c|}{48} & \multicolumn{1}{c|}{{\color[HTML]{FF0000} \textbf{0.093 0.204}}} & \multicolumn{1}{c|}{0.120 0.233} & \multicolumn{1}{c|}{0.191 0.308} & \multicolumn{1}{c|}{0.199 0.383} & \multicolumn{1}{c|}{0.229 0.339} & \multicolumn{1}{c|}{0.136 0.250} & \multicolumn{1}{c|}{0.355 0.437} & \multicolumn{1}{c|}{0.205 0.326} & \multicolumn{1}{c|}{{\color[HTML]{00CCFF} {\ul 0.099}} {\color[HTML]{00CCFF} {\ul 0.211}}} & \multicolumn{1}{c|}{0.270 0.368} & 0.646 0.610 \\ \midrule
\multicolumn{1}{c|}{} & \multicolumn{1}{c|}{12} & \multicolumn{1}{c|}{{\color[HTML]{FF0000} \textbf{0.055 0.154}}} & \multicolumn{1}{c|}{{\color[HTML]{00CCFF} {\ul 0.067}} {\color[HTML]{00CCFF} {\ul  0.165}}} & \multicolumn{1}{c|}{0.137 0.256} & \multicolumn{1}{c|}{0.151 0.322} & \multicolumn{1}{c|}{0.095 0.207} & \multicolumn{1}{c|}{0.082 0.181} & \multicolumn{1}{c|}{0.115 0.242} & \multicolumn{1}{c|}{0.094 0.200} & \multicolumn{1}{c|}{0.068 0.171} & \multicolumn{1}{c|}{0.109 0.225} & 0.199 0.336 \\
\multicolumn{1}{c|}{} & \multicolumn{1}{c|}{24} & \multicolumn{1}{c|}{{\color[HTML]{FF0000} \textbf{0.065 0.172}}} & \multicolumn{1}{c|}{{\color[HTML]{00CCFF} {\ul 0.088}} {\color[HTML]{00CCFF} {\ul  0.190}}} & \multicolumn{1}{c|}{0.111 0.225} & \multicolumn{1}{c|}{0.169 0.348} & \multicolumn{1}{c|}{0.150 0.262} & \multicolumn{1}{c|}{0.101 0.204} & \multicolumn{1}{c|}{0.210 0.329} & \multicolumn{1}{c|}{0.139 0.247} & \multicolumn{1}{c|}{0.119 0.225} & \multicolumn{1}{c|}{0.125 0.244} & 0.323 0.420 \\
\multicolumn{1}{c|}{\multirow{-3}{*}{PEMS07}} & \multicolumn{1}{c|}{48} & \multicolumn{1}{c|}{{\color[HTML]{FF0000} \textbf{0.107 0.204}}} & \multicolumn{1}{c|}{{\color[HTML]{00CCFF} {\ul 0.110}} {\color[HTML]{00CCFF} {\ul  0.215}}} & \multicolumn{1}{c|}{0.137 0.221} & \multicolumn{1}{c|}{0.196 0.384} & \multicolumn{1}{c|}{0.253 0.340} & \multicolumn{1}{c|}{0.134 0.238} & \multicolumn{1}{c|}{0.398 0.458} & \multicolumn{1}{c|}{0.311 0.369} & \multicolumn{1}{c|}{0.149 0.237} & \multicolumn{1}{c|}{0.165 0.288} & 0.390 0.470 \\ \midrule
\multicolumn{1}{c|}{} & \multicolumn{1}{c|}{12} & \multicolumn{1}{c|}{{\color[HTML]{FF0000} \textbf{0.063 0.165}}} & \multicolumn{1}{c|}{{\color[HTML]{00CCFF} {\ul 0.079}} {\color[HTML]{00CCFF} {\ul  0.182}}} & \multicolumn{1}{c|}{0.113 0.209} & \multicolumn{1}{c|}{0.162 0.337} & \multicolumn{1}{c|}{0.168 0.232} & \multicolumn{1}{c|}{0.112 0.212} & \multicolumn{1}{c|}{0.154 0.276} & \multicolumn{1}{c|}{0.165 0.214} & \multicolumn{1}{c|}{0.087 0.184} & \multicolumn{1}{c|}{0.173 0.273} & 0.436 0.485 \\
\multicolumn{1}{c|}{} & \multicolumn{1}{c|}{24} & \multicolumn{1}{c|}{{\color[HTML]{FF0000} \textbf{0.109}} 0.229} & \multicolumn{1}{c|}{{\color[HTML]{00CCFF} {\ul 0.115}} {\color[HTML]{FF0000} \textbf{0.219}}} & \multicolumn{1}{c|}{0.230 0.248} & \multicolumn{1}{c|}{0.181 0.364} & \multicolumn{1}{c|}{0.224 0.281} & \multicolumn{1}{c|}{0.141 0.238} & \multicolumn{1}{c|}{0.248 0.353} & \multicolumn{1}{c|}{0.215 0.260} & \multicolumn{1}{c|}{0.122 \color[HTML]{00CCFF} {\ul 0.221}} & \multicolumn{1}{c|}{0.210 0.301} & 0.467 0.502 \\
\multicolumn{1}{c|}{\multirow{-3}{*}{PEMS08}} & \multicolumn{1}{c|}{48} & \multicolumn{1}{c|}{{\color[HTML]{FF0000} \textbf{0.159}} \color[HTML]{00CCFF} {\ul 0.238}} & \multicolumn{1}{c|}{{\color[HTML]{00CCFF} {\ul 0.186}} {\color[HTML]{FF0000} \textbf{0.235}}} & \multicolumn{1}{c|}{0.317 0.324} & \multicolumn{1}{c|}{0.224 0.422} & \multicolumn{1}{c|}{0.321 0.354} & \multicolumn{1}{c|}{0.198 0.283} & \multicolumn{1}{c|}{0.440 0.470} & \multicolumn{1}{c|}{0.315 0.355} & \multicolumn{1}{c|}{0.189 0.270} & \multicolumn{1}{c|}{0.320 0.394} & 0.966 0.733 \\ \bottomrule
\label{tab:addlabel4}
\end{tabular}}
\end{table}

\textbf{Ultra-Long-Range Time Series Forecasting Under Multivariate Settings.}
We conduct ultra-long-range time series forecasting by taking fixed input length ($T= 96$) to predict ultra-long lengths ($H=\{1080, 1440, 1800, 2160\}$). We run all results by ourselves. Table \ref{tab:addlabel5} summarizes the results of ultra-long-range time series forecasting under multivariate settings, where - - indicates that the method fails to produce any results on that prediction length due to the out-of-memory problems. We can see from Table \ref{tab:addlabel5} that Ada-MSHyper achieves SOTA results on almost all datasets. Specifically, Ada-MSHyper gives an average error reduction of 4.97\% and 2.21\% compared to the best baseline in terms of MSE and MAE, respectively.

\begin{table}[htbp]
  \centering
  \begin{tiny}
  \caption{Full results of ultra-long-range time series forecasting results under multivariate settings.}
  \label{tab:addlabel5}
  \resizebox{\linewidth}{!}{
  \begin{tabular}{ccccccccccccc}
    \toprule
    \multicolumn{2}{c}{Models} &  \begin{tabular}[c]{@{}c@{}} Ada-MSHyper \\      (Ours)\end{tabular}& \begin{tabular}[c]{@{}c@{}}iTransformer*\\      (2024)\end{tabular} & \begin{tabular}[c]{@{}c@{}}MSHyper*\\      (2024)\end{tabular} & \begin{tabular}[c]{@{}c@{}}TimeMixer*\\      (2024)\end{tabular} & \begin{tabular}[c]{@{}c@{}}WITRAN*\\      (2023)\end{tabular} & \begin{tabular}[c]{@{}c@{}}PatchTST*\\      (2023)\end{tabular} & \begin{tabular}[c]{@{}c@{}}Dlinear*\\      (2023)\end{tabular} & \begin{tabular}[c]{@{}c@{}}Crossformer*\\      (2023)\end{tabular} & \begin{tabular}[c]{@{}c@{}}FEDformer*\\      (2022)\end{tabular} & \begin{tabular}[c]{@{}c@{}}Pyraformer*\\      (2022)\end{tabular} & \begin{tabular}[c]{@{}c@{}}Autoformer*\\      (2021)\end{tabular} \\ \midrule
    \multicolumn{2}{c|}{Metric} & \multicolumn{1}{c|}{MSE MAE} &\multicolumn{1}{c|}{ MSE MAE} & \multicolumn{1}{c|}{MSE MAE} & \multicolumn{1}{c|}{MSE MAE} & \multicolumn{1}{c|}{MSE MAE} & \multicolumn{1}{c|}{MSE MAE} & \multicolumn{1}{c|}{MSE MAE} & \multicolumn{1}{c|}{MSE MAE} & \multicolumn{1}{c|}{MSE MAE} & \multicolumn{1}{c|}{MSE MAE} & \multicolumn{1}{c}{MSE MAE} \\ \midrule
    \multicolumn{1}{c|}{} & \multicolumn{1}{c|}{1080} & \multicolumn{1}{c|}{{\color[HTML]{FF0000} \textbf{0.534 0.509}}} & \multicolumn{1}{c|}{0.562 0.521} & \multicolumn{1}{c|}{0.557 0.517} & \multicolumn{1}{c|}{0.682 0.569} & \multicolumn{1}{c|}{0.602 0.660} & \multicolumn{1}{c|}{{\color[HTML]{00B0F0} \ul 0.549} {\color[HTML]{00B0F0} \ul 0.512}} & \multicolumn{1}{c|}{0.593 0.564} & \multicolumn{1}{c|}{0.877 1.204} & \multicolumn{1}{c|}{0.699 0.615} & \multicolumn{1}{c|}{1.015 0.798} & \multicolumn{1}{c}{0.695 0.626} \\
    \multicolumn{1}{c|}{} & \multicolumn{1}{c|}{1440} & \multicolumn{1}{c|}{{\color[HTML]{FF0000} {\textbf{0.616 0.498}}}} & \multicolumn{1}{c|}{0.620 0.556} & \multicolumn{1}{c|}{0.667 0.578} & \multicolumn{1}{c|}{0.793 0.625} & \multicolumn{1}{c|}{0.705 0.878} & \multicolumn{1}{c|}{{\color[HTML]{00B0F0} {\ul 0.619} \color[HTML]{00B0F0} \ul 0.553}} & \multicolumn{1}{c|}{0.661 0.607} & \multicolumn{1}{c|}{0.863 1.175} &  \multicolumn{1}{c|}{0.621 0.567} & \multicolumn{1}{c|}{1.075 0.833} & \multicolumn{1}{c}{0.876 0.696} \\
    \multicolumn{1}{c|}{} & \multicolumn{1}{c|}{1800} & \multicolumn{1}{c|}{{\color[HTML]{FF0000} \textbf{0.689}} 0.627} &  \multicolumn{1}{c|}{0.780 0.631} & \multicolumn{1}{c|}{0.758 {\color[HTML]{00B0F0} \ul 0.624}} & \multicolumn{1}{c|}{0.877 0.643} & \multicolumn{1}{c|}{0.775 {\color[HTML]{FF0000} \textbf{0.623}}} & \multicolumn{1}{c|}{0.775 {\color[HTML]{FF0000} \textbf{0.623}}} & \multicolumn{1}{c|}{{\color[HTML]{00B0F0} {\ul 0.746}} 0.658} & \multicolumn{1}{c|}{0.849 1.163} & \multicolumn{1}{c|}{0.806 0.649} & \multicolumn{1}{c|}{1.111 0.844} & \multicolumn{1}{c}{0.852 0.704} \\
    \multicolumn{1}{c|}{\multirow{-4}{*}{ETTh1}} & \multicolumn{1}{c|}{2160} & \multicolumn{1}{c|}{{\color[HTML]{FF0000} {\textbf{0.779 0.635}}}} & \multicolumn{1}{c|}{1.102 0.736} & \multicolumn{1}{c|}{0.998 0.721} & \multicolumn{1}{c|}{1.007 0.686} & \multicolumn{1}{c|}{0.852 1.171} & \multicolumn{1}{c|}{0.851  {\color[HTML]{00B0F0} {\ul 0.665}}} &  \multicolumn{1}{c|}{{\color[HTML]{00B0F0} {\ul 0.783}} 0.667} & \multicolumn{1}{c|}{1.095 0.821} & \multicolumn{1}{c|}{0.935 0.717} & \multicolumn{1}{c|}{1.129 0.847} & \multicolumn{1}{c}{-- --} \\ \midrule
    \multicolumn{1}{c|}{} & \multicolumn{1}{c|}{1080} & \multicolumn{1}{c|}{{\color[HTML]{FF0000} \textbf{0.426 0.461}}} & \multicolumn{1}{c|}{0.486 0.488} & \multicolumn{1}{c|}{0.464 0.469} & \multicolumn{1}{c|}{0.483 0.480} & \multicolumn{1}{c|}{{\color[HTML]{00B0F0} {\ul 0.432}} 0.474} & \multicolumn{1}{c|}{0.453 {\color[HTML]{00B0F0} {\ul 0.468}}} & \multicolumn{1}{c|}{0.730 0.617} & \multicolumn{1}{c|}{1.481 0.918} & \multicolumn{1}{c|}{0.514 0.526} & \multicolumn{1}{c|}{3.224 1.458} & \multicolumn{1}{c}{0.559 0.547} \\
    \multicolumn{1}{c|}{} & \multicolumn{1}{c|}{1440} & \multicolumn{1}{c|}{{\color[HTML]{FF0000} \textbf{0.465 0.437}}} & \multicolumn{1}{c|}{0.512 0.507} & \multicolumn{1}{c|}{0.524 0.506} & \multicolumn{1}{c|}{0.547 0.510} & \multicolumn{1}{c|}{{\color[HTML]{00B0F0} {\ul 0.472 0.443}}} & \multicolumn{1}{c|}{0.513 0.501} & \multicolumn{1}{c|}{1.144 0.770} & \multicolumn{1}{c|}{1.901 1.044} & \multicolumn{1}{c|}{0.578 0.546} & \multicolumn{1}{c|}{3.254 1.548} & \multicolumn{1}{c}{0.638 0.708} \\
    \multicolumn{1}{c|}{} & \multicolumn{1}{c|}{1800} & \multicolumn{1}{c|}{{\color[HTML]{FF0000} \textbf{0.503}} 0.505} & \multicolumn{1}{c|}{0.565 0.529} & \multicolumn{1}{c|}{0.522 {\color[HTML]{FF0000} \textbf{0.496}}} & \multicolumn{1}{c|}{0.606 0.544} & \multicolumn{1}{c|}{0.626 0.610} & \multicolumn{1}{c|}{{\color[HTML]{00B0F0} {\ul 0.517}} {\color[HTML]{00B0F0} {\ul 0.503}}} & \multicolumn{1}{c|}{1.327 0.840} & \multicolumn{1}{c|}{3.109 1.486} & \multicolumn{1}{c|}{0.645 0.584} & \multicolumn{1}{c|}{3.328 1.565} & \multicolumn{1}{c}{0.776 0.689} \\
    \multicolumn{1}{c|}{\multirow{-4}{*}{ETTh2}} & \multicolumn{1}{c|}{2160} & \multicolumn{1}{c|}{{\color[HTML]{FF0000} \textbf{0.527}} {\color[HTML]{00B0F0} {\ul 0.515}}} & \multicolumn{1}{c|}{0.600 0.546} & \multicolumn{1}{c|}{{\color[HTML]{00B0F0} {\ul 0.542}} {\color[HTML]{FF0000} \textbf{0.510}}} &  \multicolumn{1}{c|}{0.616 0.557} & \multicolumn{1}{c|}{0.657 0.619} & \multicolumn{1}{c|}{0.547 0.519} & \multicolumn{1}{c|}{1.670 0.919} & \multicolumn{1}{c|}{3.630 1.485} & \multicolumn{1}{c|}{0.762 0.639} & \multicolumn{1}{c|}{3.246 1.465} & -- -- \\ \midrule
    \multicolumn{1}{c|}{} & \multicolumn{1}{c|}{1080} & \multicolumn{1}{c|}{{\color[HTML]{FF0000} \textbf{0.460 0.445}}} &  \multicolumn{1}{c|}{0.534 0.483} & \multicolumn{1}{c|}{0.520 0.465} & \multicolumn{1}{c|}{0.502 0.465} & \multicolumn{1}{c|}{{\color[HTML]{00B0F0} {\ul 0.464}} {\color[HTML]{00B0F0} {\ul 0.459}}} & \multicolumn{1}{c|}{0.494 {\color[HTML]{00B0F0} {\ul 0.459}}} & \multicolumn{1}{c|}{0.514 0.479} & \multicolumn{1}{c|}{2.588 1.236} & \multicolumn{1}{c|}{0.513 0.499} & \multicolumn{1}{c|}{1.071 0.793} & \multicolumn{1}{c}{0.651 0.551} \\
    \multicolumn{1}{c|}{} & \multicolumn{1}{c|}{1440} & \multicolumn{1}{c|}{{\color[HTML]{FF0000} \textbf{0.473 0.449}}} &  \multicolumn{1}{c|}{0.556 0.495} & \multicolumn{1}{c|}{0.542 0.477} & \multicolumn{1}{c|}{0.523 0.488} & \multicolumn{1}{c|}{0.543  {\color[HTML]{00B0F0} {\ul 0.467}}} & \multicolumn{1}{c|}{{\color[HTML]{00B0F0} {\ul 0.508}}  {\color[HTML]{00B0F0} {\ul 0.467}}} & \multicolumn{1}{c|}{0.534 0.491} & \multicolumn{1}{c|}{2.946 1.349} & \multicolumn{1}{c|}{0.511 0.494} & \multicolumn{1}{c|}{1.136 0.834} & \multicolumn{1}{c}{0.602 0.542} \\
    \multicolumn{1}{c|}{} & \multicolumn{1}{c|}{1800} & \multicolumn{1}{c|}{{\color[HTML]{FF0000} \textbf{0.492}}  {\color[HTML]{00B0F0} {\ul 0.475}}} & \multicolumn{1}{c|}{ 0.571 0.501 }& \multicolumn{1}{c|}{ 0.564 0.490 }& \multicolumn{1}{c|}{ 0.526 0.487 }& \multicolumn{1}{c|}{ 0.550 0.497 }& \multicolumn{1}{c|}{ {\color[HTML]{00B0F0} {\ul 0.504}} {\color[HTML]{FF0000} \textbf{0.434}} }& \multicolumn{1}{c|}{ 0.556 0.507 }& \multicolumn{1}{c|}{ 4.113 1.602 }& \multicolumn{1}{c|}{ 0.514 0.496 }& \multicolumn{1}{c|}{ 1.111 0.812 }& 0.641 0.558 \\
    \multicolumn{1}{c|}{\multirow{-4}{*}{ETTm1}} & \multicolumn{1}{c|}{ 2160 } & \multicolumn{1}{c|}{ {\color[HTML]{00B0F0} {\ul {0.510}}} {\color[HTML]{00B0F0} {\ul {0.483}}} } & \multicolumn{1}{c|}{ 0.555 0.499 } & \multicolumn{1}{c|}{  0.550 0.487 } & \multicolumn{1}{c|}{ 0.542 0.491 } & \multicolumn{1}{c|}{ 0.569 {\color[HTML]{FF0000} {\textbf{0.481}}} } & \multicolumn{1}{c|}{ {\color[HTML]{FF0000} {\textbf{0.507 0.481}}} } & \multicolumn{1}{c|}{ 0.556 0.515 } & \multicolumn{1}{c|}{ 4.574 1.743 } & \multicolumn{1}{c|}{ 0.551 0.516 } & \multicolumn{1}{c|}{ 1.054 0.804 } & -- -- \\ \midrule
    \multicolumn{1}{c|}{} & \multicolumn{1}{c|}{ 1080 }&  \multicolumn{1}{c|}{ {\color[HTML]{FF0000} \textbf{0.404 0.416}} }&  \multicolumn{1}{c|}{ 0.463 0.438 }&  \multicolumn{1}{c|}{ 0.464 0.439 }&  \multicolumn{1}{c|}{ 0.450 {\color[HTML]{00B0F0} {\ul 0.432}} }&  \multicolumn{1}{c|}{ {\color[HTML]{00B0F0} {\ul 0.415}} 0.434 }&  \multicolumn{1}{c|}{ 0.449 {\color[HTML]{00B0F0} {\ul 0.432}} }&  \multicolumn{1}{c|}{ 0.559 0.519 }&  \multicolumn{1}{c|}{ 2.587 1.236 }&  \multicolumn{1}{c|}{ 0.501 0.468 }&  \multicolumn{1}{c|}{ 4.879 1.733 }&  0.527 0.489 \\
    \multicolumn{1}{c|}{} & \multicolumn{1}{c|}{ 1440 } & \multicolumn{1}{c|}{ {\color[HTML]{FF0000} \textbf{0.413 0.429}} } & \multicolumn{1}{c|}{ 0.475 0.452 } & \multicolumn{1}{c|}{ 0.475 0.449 } & \multicolumn{1}{c|}{ 0.471 0.452 } & \multicolumn{1}{c|}{ {\color[HTML]{00B0F0} {\ul 0.442}} {\color[HTML]{00B0F0} {\ul 0.442}} } & \multicolumn{1}{c|}{ 0.475 0.452 } & \multicolumn{1}{c|}{ 0.699 0.593 } & \multicolumn{1}{c|}{ 2.946 1.349 } & \multicolumn{1}{c|}{ 0.495 0.480 } & \multicolumn{1}{c|}{ 4.429 1.708 } & 0.519 0.489 \\
    \multicolumn{1}{c|}{} & \multicolumn{1}{c|}{ 1800 } & \multicolumn{1}{c|}{ {\color[HTML]{FF0000} \textbf{0.435}} {\color[HTML]{00B0F0} {\ul 0.432}} } & \multicolumn{1}{c|}{ 0.468 0.453 } & \multicolumn{1}{c|}{ {\color[HTML]{00B0F0} {\ul 0.454}} 0.449 } & \multicolumn{1}{c|}{ 0.464 0.452 } & \multicolumn{1}{c|}{ 0.479 {\color[HTML]{FF0000} \textbf{0.410}} } & \multicolumn{1}{c|}{ 0.456 0.449 } & \multicolumn{1}{c|}{ 0.721 0.612 } & \multicolumn{1}{c|}{ 4.113 1.602 } & \multicolumn{1}{c|}{ 0.477 0.474 } & \multicolumn{1}{c|}{ 4.502 1.780 } & 0.503 0.496 \\
    \multicolumn{1}{c|}{\multirow{-4}{*}{ETTm2}} & \multicolumn{1}{c|}{ 2160 } & \multicolumn{1}{c|}{ {\color[HTML]{00B0F0} {\ul 0.449}} 0.457 } & \multicolumn{1}{c|}{ 0.467 0.454 } & \multicolumn{1}{c|}{ 0.463 {\color[HTML]{00B0F0} {\ul 0.451}} } & \multicolumn{1}{c|}{ 0.473 0.459 } & \multicolumn{1}{c|}{ {\color[HTML]{FF0000} \textbf{0.447}} {\color[HTML]{FF0000} \textbf{0.449}} } & \multicolumn{1}{c|}{ 0.466 0.457 } & \multicolumn{1}{c|}{ 0.639 0.572 } & \multicolumn{1}{c|}{ 4.574 1.743 } & \multicolumn{1}{c|}{ 0.473 0.477 } & \multicolumn{1}{c|}{4.454 1.758} & -- -- \\ \bottomrule
    \end{tabular}
    }
\end{tiny}
\end{table}%

\section{Ablation Studies} \label{Ablation Studies2}
To investigate the performance of Ada-MSHyper on longer prediction lengths, we compare the forecasting results of Ada-MSHyper with those of six variations (i.e., AGL, one, PH, -w/o NC, -w/o HC, and -w/o NHC) on ETTh1 dataset. The experimental results are shown in Table \ref{tab:tablelonger}. We can observe that for longer prediction lengths, -w/o NC has smaller performance degradation than other variations. The reason may be that when the prediction length increases, the model tends to focus more on macroscopic variation interactions and diminishes its emphasis on fine-grained node constraint. In addition, Ada-MSHyper performs better than other six variations even with longer prediction length, showing the effectiveness of our AHL module and NHC mechanism.

\begin{table}[htbp]
\centering
\caption{Results of different adaptive hypergraph learning methods and constraint mechanisms.}
\label{tab:tablelonger}
\resizebox{0.8 \linewidth}{!}{
\begin{tabular}{c|cccccc|cccccc|cc}
\toprule
Variation & \multicolumn{2}{c}{AGL}       & \multicolumn{2}{c}{one}           & \multicolumn{2}{c|}{PH}            & \multicolumn{2}{c}{-w/o NC}       & \multicolumn{2}{c}{-w/o HC}       & \multicolumn{2}{c|}{-w/o NHC}      & \multicolumn{2}{c}{Ada-MSHyper}                     \\ \toprule
Metric    & \multicolumn{1}{c}{MSE} & MAE & \multicolumn{1}{c}{MSE}   & MAE   & \multicolumn{1}{c}{MSE}   & MAE   & \multicolumn{1}{c}{MSE}   & MAE   & \multicolumn{1}{c}{MSE}   & MAE   & \multicolumn{1}{c}{MSE}   & MAE   & \multicolumn{1}{c}{MSE}            & MAE            \\ \toprule
1080      & \multicolumn{1}{c}{-}   & -   & \multicolumn{1}{c}{0.685} & 0.679 & \multicolumn{1}{c}{0.640} & 0.591 & \multicolumn{1}{c}{0.539} & 0.515 & \multicolumn{1}{c}{0.574} & 0.516 & \multicolumn{1}{c}{0.597} & 0.525 & \multicolumn{1}{c}{\textbf{0.534}} & \textbf{0.509} \\ 
1440      & \multicolumn{1}{c}{-}   & -   & \multicolumn{1}{c}{0.855} & 0.857 & \multicolumn{1}{c}{0.783} & 0.673 & \multicolumn{1}{c}{0.621} & 0.503 & \multicolumn{1}{c}{0.679} & 0.568 & \multicolumn{1}{c}{0.734} & 0.585 & \multicolumn{1}{c}{\textbf{0.616}} & \textbf{0.498} \\ \bottomrule
\end{tabular}}
\end{table}

\begin{figure*}[htbp]
\includegraphics[width=5in]{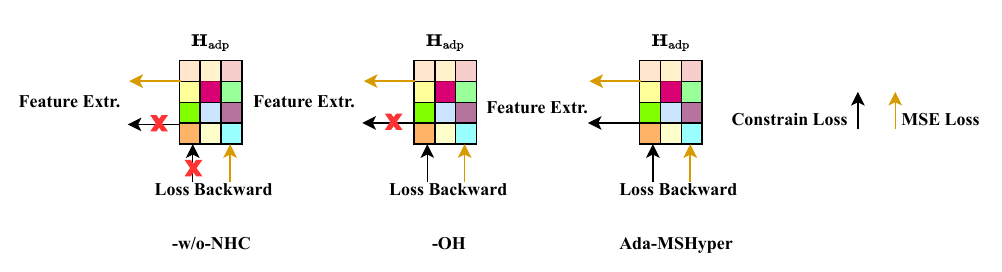}
\centering
\caption{Different optimization strategies for AHL.}
\label{addfigure4}
\end{figure*}

To investigate the impact of node and hypergraph constraints mechanism on the adaptive hypergraph learning (AHL) module, we design two variants:
(1) Removing the NHC mechanism (-w/o NHC).
(2) Only optimizing the hypergraph learning module (-OH).
We illustrate these two variants for better understanding in Figure \ref{addfigure4}.

To investigate the impact of the multi-scale feature extraction (MFE) module, we design two variants:
(1) Replacing the aggregation function in the MFE module with average pooling (-avg).
(2) Replacing the aggregation function in the MFE module with max pooling (-max).

The results for the four variants are shown in Table \ref{tab:addlabel6}. We can observe that: (1) -w/o NHC gets the worst results. This may be because lacking the constraints makes the model fail to capture implicit semantic features of the clustered nodes and learned hyperedges. (2) Compared to -OH, Ada-MSHyper yields slightly better results. The reason may be that the NHC mechanism facilitates the MFE model to more effectively aggregate similar features during multi-scale feature extraction and reduce the interference of noise. (3) -avg and -max get relatively worse performance. This may be because the lack of parameters leads to the reduction of the representative ability of Ada-MSHyper. 

\begin{table}[htbp]
\centering
\caption{Results of different AHL, MFE and multi-scale interaction methods.}
\label{tab:addlabel6}
\resizebox{0.7 \linewidth}{!}{
\begin{tabular}{@{}c|cccc|cccc|cc|cc@{}}
\toprule
Methods & \multicolumn{2}{c}{-w/o NHC} & \multicolumn{2}{c|}{-OH} & \multicolumn{2}{c}{-avg} & \multicolumn{2}{c|}{-max} & \multicolumn{2}{c|}{-r/att} & \multicolumn{2}{c}{Ada-MSHyper} \\ \midrule
Metric & MSE & MAE & MSE & MAE & MSE & MAE & MSE & MAE & MSE & MAE & MSE & MAE \\ \midrule
96 & 0.393 & 0.422 & 0.379 & 0.397 & 0.372 & 0.413 & 0.382 & 0.408 & 0.418  & 0.419 & \textbf{0.372} & \textbf{0.393} \\
336 & 0.486 & 0.459 & 0.423 & 0.439 & 0.429 & 0.440 & 0.426 & 0.437 & 0.483 & 0.454 & \textbf{0.422} & \textbf{0.433} \\
720 & 0.515 & 0.487  & 0.447 & 0.460 & 0.453 & 0.462 & 0.448 & 0.464 & 0.514 & 0.507 & \textbf{0.445} & \textbf{0.459} \\ \bottomrule
\end{tabular}}
\end{table}

To investigate the effectiveness of hypergraph convolution attention, we design one variant: Replacing the hypergraph convolution attention with the attention mechanism used in the inter-scale interaction module to update node features (-r/att). The experimental results on ETTh1 dataset are shown in Table \ref{tab:addlabel6}. We can observe that Ada-MSHyper performs better than -r/ att, which demonstrates the effectiveness of the hyperedge convolution attention used in the intra-scale interaction module.
\section{Parameter Studies} \label{Parameter Studies}
We perform parameter studies to measure the impact of the threshold $\eta$, which influences the effectiveness of the sparsity strategy. The experimental results on ETTh1 dataset are shown in Table \ref{tab:addthrehold}. We can see that the best performance can be obtained when $\eta$ is 3. The reason is that a small $\eta$ may filter out useful information and a large $\eta$ would introduce noise interference. 
\begin{table}[htbp]
\centering
\caption{Results of Ada-MSHyper with different $\eta$.}
\label{tab:addthrehold}
\resizebox{0.6 \linewidth}{!}{
\begin{tabular}{c|cccccccccc}
\toprule
Hyparameter & \multicolumn{2}{c}{$\eta$=1}      & \multicolumn{2}{c}{$\eta$=2}      & \multicolumn{2}{c}{$\eta$=3}                        & \multicolumn{2}{c}{$\eta$=4}      & \multicolumn{2}{c}{$\eta$=5}      \\ \midrule
Metric      & \multicolumn{1}{c}{MSE}   & MAE   & \multicolumn{1}{c}{MSE}   & MAE   & \multicolumn{1}{c}{MSE}            & MAE            & \multicolumn{1}{c}{MSE}   & MAE   & \multicolumn{1}{c}{MSE}   & MAE   \\ \midrule
96          & \multicolumn{1}{c}{0.407} & 0.415 & \multicolumn{1}{c}{0.390} & 0.397 & \multicolumn{1}{c}{\textbf{0.372}} & \textbf{0.393} & \multicolumn{1}{c}{0.387} & 0.396 & \multicolumn{1}{c}{0.419} & 0.418 \\ 
336         & \multicolumn{1}{c}{0.547} & 0.500 & \multicolumn{1}{c}{0.476} & 0.443 & \multicolumn{1}{c}{\textbf{0.422}} & \textbf{0.433} & \multicolumn{1}{c}{0.438} & 0.435 & \multicolumn{1}{c}{0.560} & 0.510 \\ 
720         & \multicolumn{1}{c}{0.450} & 0.463 & \multicolumn{1}{c}{0.476} & 0.465 & \multicolumn{1}{c}{\textbf{0.445}} & \textbf{0.459} & \multicolumn{1}{c}{0.460} & 0.459 & \multicolumn{1}{c}{0.473} & 0.474 \\ \bottomrule
\end{tabular}}
\end{table}
\section{Visualization}\label{case appedix}
\begin{figure*}[!t]
\centering
\subfloat[Input sequence]{\includegraphics[width=0.30\textwidth]{fig1/inputsequence1}%
\label{ad_first_case}}
\hfil
\subfloat[Ada-MSHyper-w/o NHC]{\includegraphics[width=0.30\textwidth]{fig1/NHC1}%
\label{ad_fourth_case}}
\subfloat[Ada-MSHyper-w/o NC]{\includegraphics[width=0.30\textwidth]{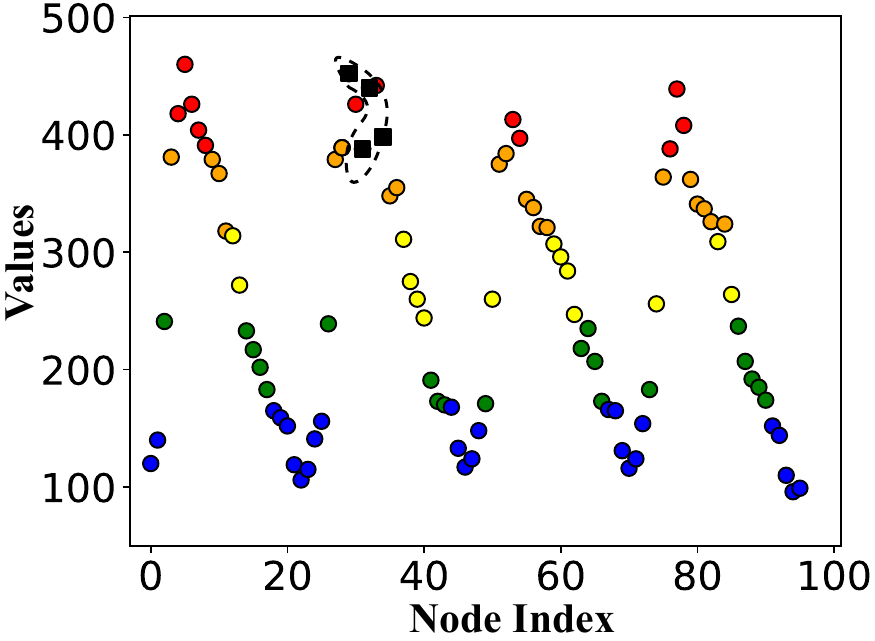}%
\label{ad_third_case}}
\hfil
\subfloat[Ada-MSHyper-w/o HC]{\includegraphics[width=0.30\textwidth]{fig1/WNHC1}%
\label{ad_second_case}}
\subfloat[Ada-MSHyper]{\includegraphics[width=0.30\textwidth]{fig1/AdaMSHy1}%
\label{ad_fifth_case}}
\caption{Visualization the node constrain effect on Electricity dataset.}
\label{addfigure1}
\end{figure*}

\textbf{Visualization of Node Constraint.} 
As shown in Figure \ref{ad_first_case}, each time step is denoted a node of the hypergraph at the finest scale. We categorize the nodes into four groups based on the node values of original inputs, and draw them using different colors. For a target node, nodes of the same color may be regarded as those sharing similar semantic information with the target node, while nodes of other colors may be regarded as noise. Then, we draw the nodes related to the target node based on the incidence matrix $\mathbf{H}^1$ of the learned hypergraph in the black color.

We random select samples at the same time step from three variants, i.e., without node constraint (-w/o NC), without hyperedge constraint (-w/o HC), and without node and hyperedges constraints (-w/o NHC), and plot these three samples with samples from original inputs and Ada-MHyper. 

We can observe that:
(1) In Figure \ref{ad_fourth_case} and Figure \ref{ad_third_case}, the related nodes of the target node are almost neighboring nodes. However, some noise, plotted as orange, is included as well. The reason may be that -w/o NHC and -w/o NC cannot distinguish noise information without node constraint, i.e., cannot consider the semantic similarity to cluster nodes. 
(2) In Figure \ref{ad_second_case} and Figure \ref{ad_fifth_case}, since -w/o HC and the proposed Ada-MSHyper have node constraint, both of them can cluster neighboring and distant but strongly correlated nodes, and they can also mitigate the interference of noise, indicating the effectiveness of node constraint.

\begin{figure*}[htbp]
\includegraphics[width=0.35\textwidth]{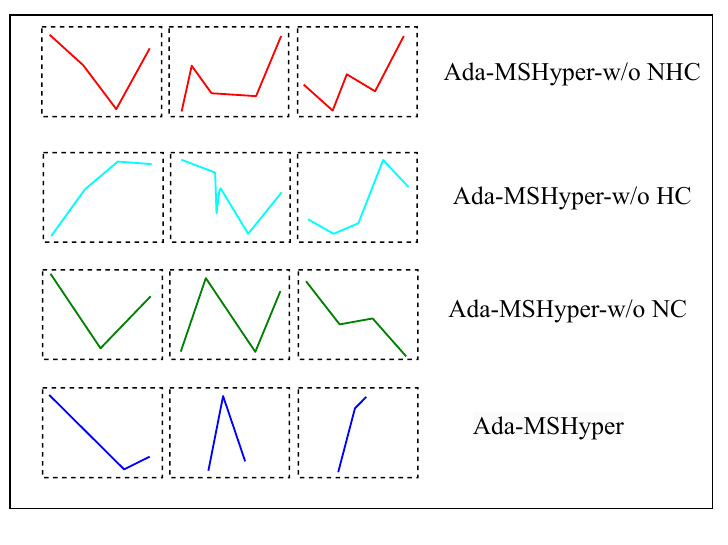}
\centering
\caption{Different temporal variations learned by different methods.}
\label{addfigure3}
\end{figure*}

\textbf{Visualization of Hyperedge Constraint.}
We use the samples which are used in the visualization of node constraint. We visualize the sequentially connecting nodes that belong to the same hyperedges whose indices are $\{4, 8, 12\}$. Figure \ref{addfigure3} shows three types of temporal variations learned by -w/o NC, -w/o HC, -w/o NHC, and Ada-MSHyper, respectively.
We can observe that: (1) -w/o HC and -w/o NHC exhibit irregular temporal variations in comparison to -w/o NC and Ada-MSHyper. The reason may be that without the hyperedge constraint, these methods are unable to adequately differentiate temporal variations entangled in temporal patterns. (2) Compared to -w/o NC, Ada-MSHyper exhibits relatively simple temporal variations. The reason may be that influenced by node constraint, the temporal variations extracted by Ada-MSHyper contain less noise.

\begin{figure*}[htbp]
\centering
\subfloat[Input sequence]{\includegraphics[width=0.35\textwidth]{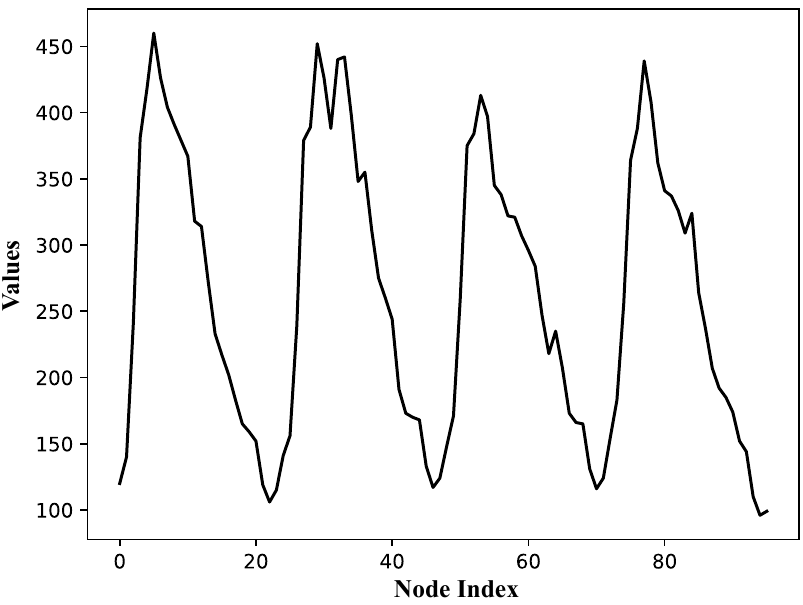}%
\label{ad_first_case1}}
\subfloat[Ada-MSHyper]{\includegraphics[width=0.35\textwidth]{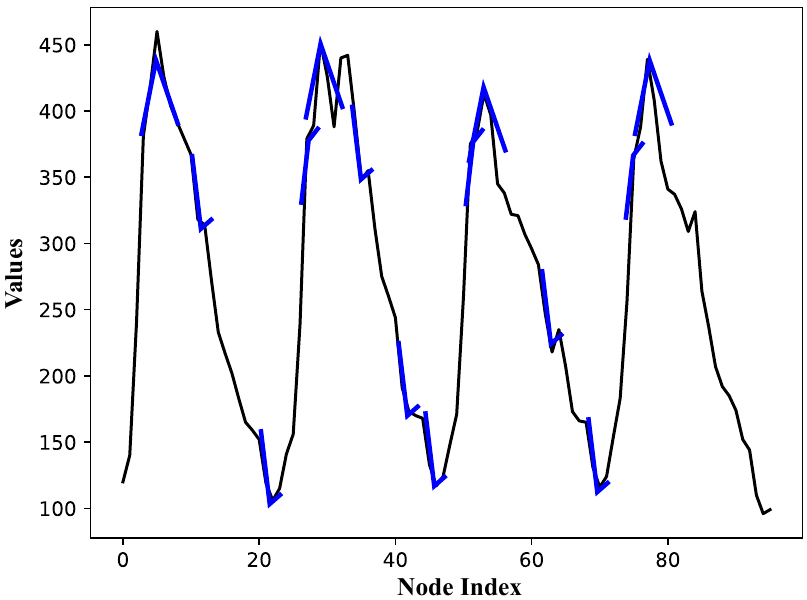}%
\label{ad_third_case1}}
\caption{Visualization the hyperedge constraint effect on Electricity dataset.}
\label{addfigure2}
\end{figure*}

We also matched the temporal variations extracted by Ada-MSHyper to the sample sequences. As shown in Figure \ref{ad_third_case1}, we can observe that these variations can represent inherent changes. We speculate that by introducing hyperedge constraint, the model will treat temporal variations with different shapes as distinct positive and negative examples. In addition, the differentiated temporal variations are like a kind of Shapelet, akin to those used in NLP and CV, enabling a better representation of temporal patterns within time series. 

\section{Limitations and Future Works} \label{limitations}
In the future, we will extend our work in the following directions. Firstly, due to our NHC mechanism can cluster nodes with similar semantic information and differentiate temporal variations within each scales, It is interesting to correlate the inherent temporal variations with corpora used in natural language processing, and leverage large language models to investigate deeper correlations between corpora and TS data. Secondly, compared to natural language processing and computer vision, time series analysis has access to fewer datasets, which may limit the expressive power of the models. Therefore, in the future, we plan to compile larger datasets to validate the generalization capabilities of our models on more extensive data.

\section{Broader Impacts}\label{Broader Impacts}
In this paper, we propose Ada-MSHyper for time series forecasting. Extensive experimental results demonstrate the effectiveness of Ada-MSHyper. Our paper mainly focuses on scientific research and has no obvious negative social impact.

\end{document}